\documentclass[10pt,twocolumn,letterpaper]{article}

\usepackage[pagenumbers]{cvpr}   

\usepackage[accsupp]{axessibility}
\usepackage{graphicx}
\usepackage{amsmath}
\usepackage{amssymb}
\usepackage{booktabs}
\usepackage{algorithm}
\usepackage{algorithmic}
\usepackage{makecell}
\usepackage{multirow}
\usepackage{bm}
\usepackage{enumitem}
\usepackage{xfrac}
\usepackage{xcolor}
\usepackage{url}

\usepackage[pagebackref,breaklinks,colorlinks]{hyperref}

\usepackage[capitalize]{cleveref}
\crefname{section}{Sec.}{Secs.}
\Crefname{section}{Section}{Sections}
\Crefname{table}{Table}{Tables}
\crefname{table}{Tab.}{Tabs.}

\newenvironment{myitemize}{%
    \begin{list}{\textbullet}{%
            \setlength{\topsep}{0pt} \setlength{\parskip}{1pt}  \setlength{\partopsep}{0pt}
            \setlength{\parsep}{0pt}  \setlength{\itemsep}{1pt}
            \setlength{\leftmargin}{5mm}}}
    {\end{list}}

\newcommand{\PreserveBackslash}[1]{\let\temp=\\#1\let\\=\temp}
\newcolumntype{C}[1]{>{\PreserveBackslash\centering}p{#1}}
\newcolumntype{R}[1]{>{\PreserveBackslash\raggedleft}p{#1}}
\newcolumntype{L}[1]{>{\PreserveBackslash\raggedright}p{#1}}

\def\cvprPaperID{3921} 
\def\confName{CVPR}
\def\confYear{2022}

\begin{document}

\title{Blind Image Super-resolution with Elaborate Degradation Modeling on \\ Noise and Kernel}
\author{Zongsheng Yue$^{1}$, Qian Zhao$^{2}$, Jianwen Xie$^{3}$, Lei Zhang$^{4}$, Deyu Meng$^{2,5}$, Kwan-Yee K. Wong$^{1}$ \vspace{2mm} \\ 
$^{1}$The University of Hong Kong, Hong Kong, China ~ $^{2}$Xi'an Jiaotong University, Xi'an, China \\
$^{3}$Cognitive Computing Lab, Baidu Research, Bellevue, USA \\
$^{4}$The Hong Kong Polytechnic University, Hong Kong, China\\
$^{5}$Peng Cheng Laboratory, Shenzhen, China\\
}

\maketitle

\begin{abstract}
    While researches on model-based blind single image super-resolution (SISR) have achieved tremendous successes recently,
    most of them do not consider the image degradation sufficiently. Firstly, they always assume image noise obeys an
    independent and identically distributed (i.i.d.) Gaussian or Laplacian distribution, which largely underestimates
    the complexity of real noise. Secondly, previous commonly-used kernel priors (e.g., normalization, sparsity) are not
    effective enough to guarantee a rational kernel solution, and thus degenerates the performance of subsequent SISR task.
    To address the above issues, this paper proposes a model-based blind SISR method under the probabilistic framework,
    which elaborately models image degradation from the perspectives of noise and blur kernel. Specifically, instead of the
    traditional i.i.d. noise assumption, a patch-based non-i.i.d. noise model is proposed to tackle the complicated real
    noise, expecting to increase the degrees of freedom of the model for noise representation. As for the blur kernel,
    we novelly construct a concise yet effective kernel generator, and plug it into the proposed blind SISR method as an
    explicit kernel prior (EKP).  To solve the proposed model, a theoretically grounded Monte Carlo EM algorithm is
    specifically designed.  Comprehensive experiments demonstrate the superiority of our method over current
    state-of-the-arts on synthetic and real datasets. The source code is available at \url{https://github.com/zsyOAOA/BSRDM}.
\end{abstract}

\section{Introduction} \label{sec:intro}
Single image super-resolution (SISR) is a fundamental problem in computer vision. It aims to
recover the sharp detailed high-resolution (HR) counterpart from an observed low-resolution (LR) image.
Image degradation, the functional opposite of image super-resolution, is the process of generating
a LR image from the HR one.
Unfortunately, the degradation model is always unknown while complicated, making the blind SISR problem
extremely challenging. How to rationally and practically model the degradation is therefore of great significance
in blind SISR.

\begin{figure*}[t]
    \centering
    \includegraphics[scale=0.58]{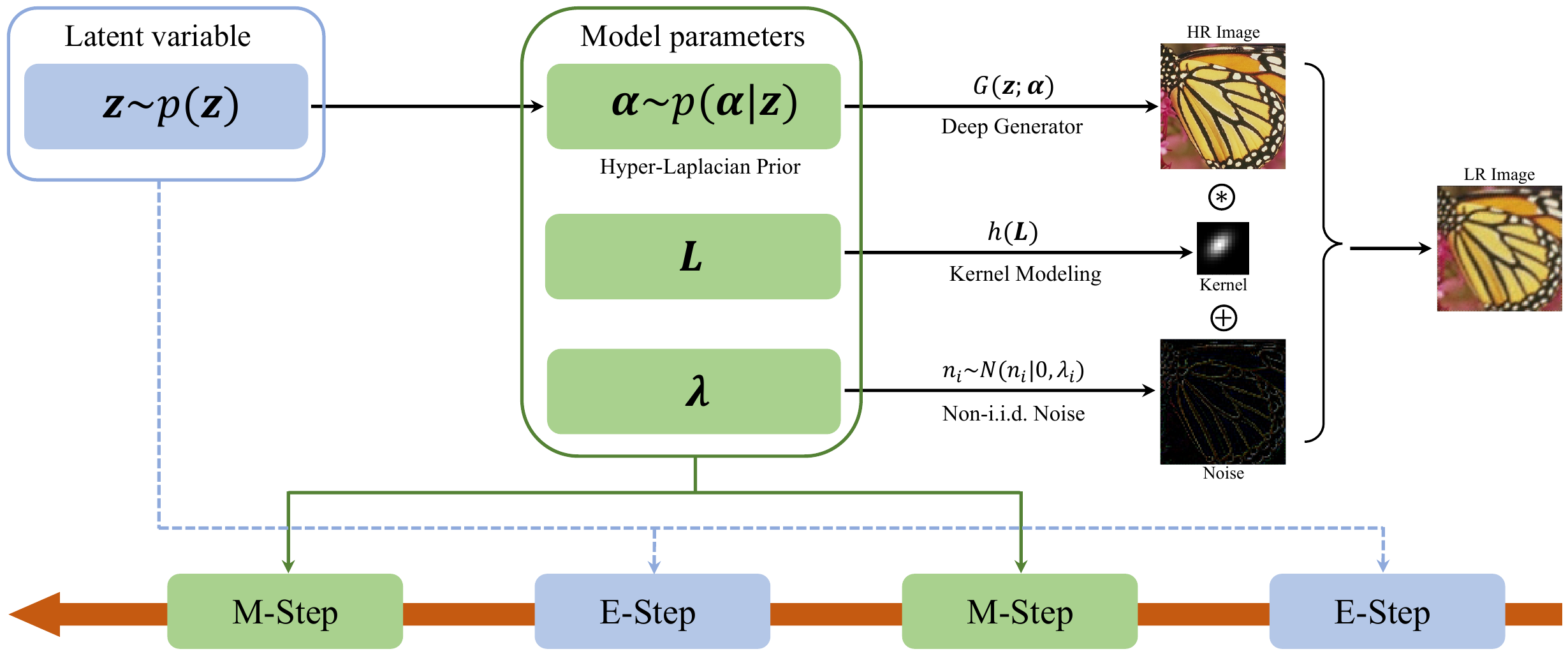}
    \vspace{-3mm}
    \caption{\footnotesize An overview of the proposed SISR method and the corresponding EM algorithm. A probabilistic model is constructed to depict the generation process of the observed
        LR image, which mainly involves two groups of parameters, including the latent variable $\bm{z}$ and the model parameters $\{\bm{\alpha},\bm{L},\bm{\lambda}\}$.
        A Monte Carlo EM algorithm is designed to alternately update them in the E-Step and M-Step, respectively.}
    \label{fig:framework}
    \vspace{-3mm}
\end{figure*}

Early methods \cite{hou1978cubic,thevenaz2000image,li2001new,timofte2014a+} simply regard SISR as an interpolation problem.
They have fast processing speed but always blur high frequency details. Later methods begin to consider the image
degradation, and can be roughly divided into two categories, namely model-based methods and learning-based methods.
From the Bayesian perspective, model-based methods~\cite{russell2003exploiting,sun2008image,kim2010single,
gu2015convolutional,ren2020neural,liang2021flow} firstly build a generative model based on the image degradation and
then estimate the blur kernel and the HR image under the maximum a posteriori (MAP) framework. Such MAP estimation is
implemented for each LR image individually, and thus tends to achieve better generalization for unknown degradations.
Learning-based methods~\cite{gu2019blind,luo2020,yue2020variational,liang2021flow}, on the other hand, aim to learn a
unified super-resolver based on a large amount of LR/HR image pairs synthesized according to the pre-assumed degradation
model. Recently, to improve their generalization,
some works~\cite{yuan2018unsupervised,fritsche2019frequency,ji2020real,maeda2020unpaired,wolf2021deflow}
attempt to learn the degradation model from unpaired real image data. However, these learning-based methods rely heavily
on the collected training data, and may suffer from a severe performance drop when unseen degradations show up in testing.
In this paper, we follow the model-based methodology for its better generalization capability.

Most of the model-based blind SISR methods can be generally formulated as the following MAP problem: 
\begin{equation}
    \max_{\bm{x}, \bm{k}} \log p(\bm{y}|\bm{x}, \bm{k}) + \log p(\bm{k}) + \log p(\bm{x}),
    \label{eq:model_based_SISR_MAP}
\end{equation}
where $\bm{y}$, $\bm{x}$, and $\bm{k}$ denote the observed LR image, the underlying HR image, and the blur kernel, respectively. The last term
represents the image prior, while the first and second terms deliver our knowledges on the degradation model (i.e., noise distribution and kernel prior).
Most of the existing methods focus on designing more rational image priors, such as gradient profile~\cite{sun2008image}, sparsity~\cite{dong2012nonlocally,kim2010single},
DIP~\cite{ulyanov2018deep} and so on~\cite{rudin1992nonlinear,gu2015convolutional,krishnan2009fast,glasner2009super,michaeli2013nonparametric,pan2020exploiting}. However,
they often do not sufficiently consider the degradation model:
\begin{myitemize}
    \item As for noise modeling, most of current method adopt the independent and identically distributed (i.i.d.) Gaussian or Laplacian distribution to model
        the noise. 
        Such a simple noise assumption, however, usually underestimates the complexity of real image noise and shows limited robustness in practical
        applications. For example, the most common camera sensor noise affected by the in-camera pipeline is signal-depedent, and thus
        exhibits evident non-i.i.d. property in statistics.
    \item As for kernel modeling, traditional methods often ignore it or adopt some heuristic priors, e.g., normalization (i.e., the kernel
        elements sum to $1$)~\cite{jin2018normalized} and sparsity~\cite{BellKligler2019}, which usually cannot guarantee a rational kernel solution.
        Recently, Liang \textit{et al.}~\cite{liang2021flow} trained an implicit mapping parameterized as a convolutional neural network (CNN)
        from the latent noises to anisotropic Gaussian kernels, and then embedded it into the blind SISR as a kernel prior. Albeit achieving evident
        performance improvement, this method depends on a time-consuming and labor-cubersome pre-training phase. Moreover, the fitting error, which is
        inevitable during training, may be enlarged in the alternate iterations between the kernel estimation and super-resolution tasks.
        The performance of blind SISR can be therefore further improved by designing an explicit yet effective kernel prior.
\end{myitemize}

To address the above issues, this paper proposes a probabilistic blind SISR method that elaborately considers the noise and kernel
modeling (see Fig.~\ref{fig:framework}). To better model the complicated real noise, a patch-based non-i.i.d. Gaussian noise assumption is adopted
instead of the conventional i.i.d. one. Under such setting, each $p \times p$ image patch has its own noise parameter,
which complies better with the configurations of real noise. As for blur kernel, we observe that it can be formulated
as an explicit and differentiable function in terms of the covariance matrix. This inspire us to construct an explicit
kernel prior (EKP) for the generally used anisotropic Gaussian kernel, which can be easily embedded into current deep
learning (DL)-based blind SISR methods. In summary, the contributions of this work is three-fold:
\begin{enumerate}[topsep=0pt, partopsep=0pt, parsep=0pt, itemsep=3pt, leftmargin=5mm]
    \item Different from the commonly-used i.i.d. Gaussian or Laplacian distribution, a patch-based non-i.i.d. noise
        distribution is employed in the proposed method, making it able to handle complicated real noise.
    \item A generative kernel prior named EKP is novelly constructed for the blind SISR task. It is with explicit and
        concise form, and substantiated to be able to attain a more stable kernel estimation for SISR.
    \item A theoretically grounded Monte Carlo EM algorithm (see Fig.~\ref{fig:framework}) is designed to solve our
        proposed model.
\end{enumerate}

\section{Related Work}
In this section, we briefly review the literatures on image degradation models and blind SISR.

\subsection{Image Degradation Model}
Image degradation model is a long-standing and open research topic in SISR. The most common and also simplest degradation model is bicubic downsampling, which is widely used
to synthesize training and testing data in many SISR works~\cite{gu2015convolutional,dong2014learning,kim2016accurate,lai2017deep,zhang2018image}.
More general degradation models, consisting of a sequence of blurring, downsampling and noise addition, are also widely adopted by previous
works~\cite{riegler2015conditioned,zhang2018learning,shocher2018zero,luo2020,zhang2020deep}. Recently, Zhang \textit{et al.}~\cite{zhang2021designing} propose a more practical
degradation model by introducing a random shuffle strategy among the blurring, downsampling and noise addition,  and more practical camera sensor and JPEG compression noises
in the noise addition procedure. Furthermore, Wang \textit{et al.}~\cite{wang2021real} consider the common ringing and overshoot artifacts, and propose a high-order
degradation model to cover a larger degradation space.

\subsection{Blind SISR Methods} \label{subsec:degradation}
As mentioned in the introduction, other than the heuristic interpolation-oriented
methods~\cite{hou1978cubic,keys1981cubic,li2001new,thevenaz2000image},
most of the existing methods can be loosely divided into two categories, namely model-based and learning-based methods. Even though this paper focuses on model-based methods,
we also briefly review learning-based methods for completeness.

\noindent{\textbf{Model-based Methods}}.
Model-based methods mainly focus on designing the three terms in Eq.~\eqref{eq:model_based_SISR_MAP}, i.e, the likelihood, kernel prior, and image prior.
The image prior has received more attentions in the past decades. Typical traditional image priors include total variation (TV)~\cite{rudin1992nonlinear},
hyper-Laplacian~\cite{krishnan2009fast}, gradient profile~\cite{sun2008image}, sparsity~\cite{kim2010single}, and non-local similarity~\cite{dong2012nonlocally}.
With the prevalence of deep learning, more DL-base image priors have been proposed. A representative work is proposed by Ulyanov \textit{et al.}~\cite{ulyanov2018deep},
namely deep image prior (DIP), to capture the low-level image statistics. Shocher \textit{et al.}~\cite{shocher2018zero} attempt to recover the HR image
using the prior of patch recurrence across scales. More related works can be found in~\cite{pan2020exploiting,ArjomandBigdeli2017}.

Kernel prior is another important part in the MAP framework. Traditional blind SISR methods only consider some heuristic kernel priors, such as
normalization~\cite{jin2018normalized} and sparsity~\cite{BellKligler2019}.
Recently, some works begin to implicitly model the kernel by DNN. For example, Ren \textit{et al.}~\cite{ren2020neural} propose to model the
kernel prior using a multilayer perceptron (MLP), while Liang \textit{et al.}~\cite{liang2021flow} train a flow-based kernel prior named FKP for
blind SISR. Instead of such an implicitly modeling manner, this paper attempts to design an explicit and concise kernel prior, hoping to induce
a more stable kernel estimation in blind SISR task.

As for the likelihood, most of the existing methods adopt an i.i.d. Gaussian or Laplacian distribution, which often fails to comply with
the configurations of real noise and causes a performance drop in real scenarios.
To address this issue, this work employs a non-i.i.d. noise modeling method to better deliver the real noise configurations and thus improve
its generalized capability.

\noindent\textbf{Learning-based Methods}. The main idea of learning-based methods is to learn a super-resolver from large amount of pre-simulated LR/HR image pairs.
Dong \textit{et al.}~\cite{dong2014learning} firstly propose to learn an end-to-end CNN mapping from LR to HR images. Later, plenty of CNN architectures are
designed for SISR~\cite{kim2016accurate,tai2017memnet,zhang2018image,wang2018recovering,he2019ode,liu2020residual,mei2020image}.
Recently, a flurry of unpaired SISR methods~\cite{yuan2018unsupervised,fritsche2019frequency,ji2020real,maeda2020unpaired,wolf2021deflow} have been proposed, due to
the fact that real LR images rarely come with the corresponding HR images in practice.

\section{The Proposed Method}
\subsection{Degradation Assumption}
Various degradation models have been proposed in previous works. Most of them can be written as a downsampling with a subsequent 
noise addition process, i.e.,
\begin{equation}
    \bm{y} = D(\bm{x};\bm{k}, \downarrow_s) + \bm{n}, 
    \label{eq:degradation_generation_V1}
\end{equation}
where $\bm{y}$ and $\bm{x}$ denote the LR and HR images, respectively, $D(\bm{x};\bm{k}, \downarrow_s)$ represents the downsampling process with a blur
kernel $\bm{k}$ and $s$-fold downsampler $\downarrow_s$, and $\bm{n}$ is the noise.
In fact, the real LR image may be also obtained by firstly adding noise and then downsampling the HR image~\cite{zhang2021designing}, which makes the noise 
more complicated. This process can also be formulated in the same format as Eq.~\eqref{eq:degradation_generation_V1}, i.e.,
\begin{equation}
    \bm{y} = D(\bm{x}+\bm{n};\bm{k}, \downarrow_s) = D(\bm{x};\bm{k}, \downarrow_s) + \hat{\bm{n}}, 
    \label{eq:degradation_generation_V2}
\end{equation}
where $\hat{\bm{n}}=D(\bm{n};\bm{k}, \downarrow_s)$. Hence, we only need to consider the degradation sequence in Eq.~\eqref{eq:degradation_generation_V1}.

For the blur kernel $\bm{k}$, we assume it to be the general anisotropic Gaussian kernel which is sufficient for SISR as pointed out in~\cite{riegler2015conditioned,zhang2020deep}.
Furthermore, considering different settings for the downsampler $\downarrow_s$ (e.g., bicubic~\cite{kim2016accurate} and
direct\footnote{Direct downsampler with a scale factor $s$ means keeping the upper-left pixel for each distinct $s \times s$ patch and discarding the rest.}~\cite{zhang2020deep})
and the imposed order between the blurring and downsampling procedures (i.e., ($\bm{x}* \bm{k}) \downarrow_s$~\cite{zhang2020deep} and $(\bm{x}\downarrow_s)* \bm{k}$~\cite{zhang2019deep},
where $*$ is the convolution operator), we can obtain multiple different degradation assumptions based on Eq.~\eqref{eq:degradation_generation_V1}.
This paper aims to propose a blind SISR method with elaborate considerations on noise and kernel modeling, which does not depend on the format of the downsampler and the specific
imposed order between the blurring and downsampling procedures. For the ease of presentation, we adopt the most widely used degradation assumption to
construct our SISR model in the next subsection, i.e.,
\begin{equation}
    \bm{y} = (\bm{x} * \bm{k}) \downarrow_s^d + \bm{n}, 
    \label{eq:degradation_generation}
\end{equation}
where $\downarrow_s^d$ is the direct downsampler with a scale factor $s$.

\subsection{Probabilistic SISR Model} \label{subsec:model}
In this subsection, we are going to build our blind SISR method based on the degradation model in Eq.~\eqref{eq:degradation_generation},

\vspace{1mm}\noindent\textbf{Non-i.i.d. Noise Modeling}.
Different from the traditional i.i.d. Gaussian or Laplacian noise assumption on the whole image, a patch-based non-i.i.d. noise model is proposed in this work.
Given any observed LR image $\bm{y}\in \mathbb{R}^{h\times w}$, where $h$ and $w$ denote the image height and width, respectively, we regard $\bm{y}$ as $N$ ($N=hw$) highly
overlapped $p \times p$ patches. Furthermore, we assume that the noises contained in each patch obey a different zero-mean Gaussian distribution with its
own variance parameter. Specifically, considering the $i$-th image patch centered at $y_i$, we have
\begin{equation}
y_i \sim \mathcal{N}\left(y_i|[(\bm{x}*\bm{k})\downarrow_s^d]_i, \lambda_i\right), ~i=1,2,\cdots,N,
    \label{eq:noise_likelihood}
\end{equation}
where $\lambda_i$ is the noise variance for the $i$-th image patch.

In previous researches, they often assume the noise as additive white Gaussian noise (AWGN), which
is indeed a special case of our non-i.i.d. noise distribution. By regarding the whole image as one large patch with size $h \times w$, our noise model then naturally
degenerates to AWGN, but with noise variance parameter being automatically updated during learning (see Sec.~\ref{sec:em_alg}) instead of manually adjusted.

\vspace{1mm}\noindent\textbf{Kernel Prior}.
Based on the anisotropic Gaussian assumption on the blur kernel, we construct a concise yet effective kernel prior.
For any blur kernel $\bm{k}$ with size $(2r+1) \times (2r+1)$, it is defined as follows:
\begin{equation}
    k_{ij} = \frac{1}{2\pi} \sqrt{\left\vert\bm{\Lambda}\right\vert} \exp \left\{ -\frac{1}{2}\bm{S}^T\bm{\Lambda}\bm{S} \right\},  ~ ~ i,j \in \{-r, \cdots, r\}, 
    \label{eq:kernel_gaussian}
\end{equation}
where $\bm{\Lambda}$ is the precision matrix, $\bm{S}=\left[\begin{smallmatrix} i \\ j \end{smallmatrix}\right]$ is the spatial coordinate.
From Eq.~\eqref{eq:kernel_gaussian}, it can be observed that the blur kernel is completely determined by the precision matrix $\bm{\Lambda}$
after fixing the kernel size. Note that Eq.~\eqref{eq:kernel_gaussian} is differentiable w.r.t. $\bm{\Lambda}$. This implies
that it can be regarded as a kernel generator, in which $\bm{\Lambda}$ can be easily optimized with stochastic gradient descent (SGD)
under the DL framework.

Another tricky issue is how to guarantee the positive-definiteness of the precision matrix $\bm{\Lambda}$ during optimization. Inspired by
the Cholesky decomposition, we reparameterize $\bm{\Lambda}$ as follows:
\begin{equation}
    \bm{\Lambda} = \bm{L}\bm{L}^T,
    \label{eq:cholesky_decompostion}
\end{equation}
where $\bm{L} \in \mathcal{R}^{2 \times 2}$ is a lower triangular matrix. By substituting Eq.~\eqref{eq:cholesky_decompostion} into Eq.~\eqref{eq:kernel_gaussian}, we
obtain the following explicit kernel prior termed EKP,
\begin{equation}
    k_{ij} = h(\bm{L})= \frac{1}{2\pi} \left\vert\bm{L}\right\vert \exp \left\{ -\frac{1}{2}\bm{S}^T\bm{L}\bm{L}^T\bm{S} \right\}.
    \label{eq:kernel_generator}
\end{equation}
In practice, to make $\bm{L}$ be triangular during optimization, we rewrite $\bm{L}$ as $\bm{L}=\bm{Q}\odot\bm{M}$,
where $\bm{M}=\left[\begin{smallmatrix} 1 & 0 \\ 1 & 1 \end{smallmatrix}\right]$ and $\odot$ is the Hadamard product, and turn to optimize $\bm{Q}$.


To our best knowledge, the most effective kernel prior for SISR is FKP~\cite{liang2021flow}. The main idea of FKP is to firstly train a deep generator that
maps the latent noises to anisotropic Gaussian kernels, and then use the pre-trained generator to estimate the blur kernel by only adjusting the latent noises.
The inevitable fitting error of this generator may be enlarged when applying it in blind SISR,
and thus limits the final performance. Comparing with FKP, the advantages of our proposed EKP is three-fold: 1) EKP is an explicit kernel generator
that does not rely on pre-training, making it more convenient to be used in SISR. 2) The generated kernel by EKP is always an exact anisotropic Gaussian kernel, 
which naturally avoids the issue of fitting error in FKP. 3) In EKP, the kernel $\bm{k}$ is completely controlled by $\bm{L}$, which contains much fewer parameters
than that of the latent noise vector in FKP ($3$ vs. $11^2/15^2/19^2$ for scale 2/3/4, respectively). This makes EKP more easier to be optimized after pluging into
the blind SISR as a kernel prior.

\vspace{1mm}\noindent\textbf{Image Prior}.
We employ a CNN-based generator $G$ to generate the HR image from the latent space, i.e.,
\begin{equation}
    \bm{x} = G(\bm{z};\bm{\alpha}),
    \label{eq:prior_DIP}
\end{equation}
where $\bm{z}$ and $\bm{\alpha}$ denote the latent variable and network parameters, respectively.
As demonstrated in~\cite{ulyanov2018deep}, $G$ is very easy to overfit onto the image noise due to the powerful fitting capability of CNN.
Therefore, we introduce the conventional hyper-Laplacian prior to constrain the statistical regularity of the generated
HR image through the following joint distribution of $\bm{\alpha}$ and $\bm{z}$:
\begin{align}
    (\bm{\alpha}, \bm{z}) &\sim p(\bm{\alpha},\bm{z}) = p(\bm{\alpha}|\bm{z})p(\bm{z}), \label{eq:joint_prior}\\
    p(\bm{\alpha}|\bm{z}) & \propto \exp \left(
        - \rho \sum_{k=1}^{2} \left\vert f_k * G(\bm{z};\bm{\alpha})\right\vert^{\gamma} 
    \right),  \label{eq:hyper_laplacian} \\
    p(\bm{z}) &=\mathcal{N}(\bm{z}|0, \bm{I}), \label{eq:latent_prior}
\end{align}
where $\{f_k\}_{k=1}^2$ are the gradient filters along the horizontal and vertical directions, $\rho$ and $\gamma$ are both hyper-paramters.

As for the generator G, we follow the ``hourglass'' architecture in DIP~\cite{ulyanov2018deep} but
use a tiny version that contains much fewer parameters.
The detailed network architecture can be found in appendix.

\subsection{MAP Estimation}
According to Eqs.~\eqref{eq:noise_likelihood}-\eqref{eq:latent_prior}, a full probabilistic model is constructed. Under the MAP framework, our goal
turns to maximize the following posterior:
\begin{equation}
    p(\bm{\alpha},\bm{L},\bm{\lambda}|\bm{y}) \propto \int p(\bm{y}|\bm{\alpha}, \bm{L}, \bm{\lambda}, \bm{z}) p(\bm{\alpha}|\bm{z}) p(\bm{z}) \mathrm{d}\bm{z}.
    \label{eq:map_goal}
\end{equation}
Note that we have omitted the prior terms $p(\bm{L})$ and $p(\bm{\Lambda})$, since they are set as non-informative priors in our model.
Taking the logarithm of both sides of Eq.~\eqref{eq:map_goal}, we have the following maximization problem:
\begin{align}
    &\mathrel{\phantom{=}} \max_{\bm{\alpha},\bm{L},\bm{\lambda}} \log p(\bm{\alpha},\bm{L},\bm{\lambda}|\bm{y}) \notag \\
    &= \log \int p(\bm{y}|\bm{\alpha}, \bm{L}, \bm{\lambda}, \bm{z}) p(\bm{\alpha}|\bm{z}) p(\bm{z}) \mathrm{d}\bm{z} + \text{const}. 
    \label{eq:log_map_goal}
\end{align}

\section{Inference Algorithm} \label{sec:em_alg}
Inspired by~\cite{xie2019learning,xie2020motion}, we design a Monte Carlo
expectation-maximization (EM) algorithm~\cite{dempster1977maximum} to solve Eq.~\eqref{eq:log_map_goal}, which
alternately samples the latent variable $\bm{z}$ from its posterior $p(\bm{z}|\bm{y})$ in E-Step and updates
the model parameters $\{\bm{\alpha}, \bm{L}, \bm{\lambda}\}$
in M-Step. The whole inference framework is
illustrated in Fig.~\ref{fig:framework}.

\vspace{1mm}\noindent\textbf{E-Step}. Given current model parameters $\{\bm{\alpha}_{\text{old}}, \bm{L}_{\text{old}},$ $\bm{\lambda}_{\text{old}}\}$, we denote the posterior of $\bm{z}$ under
them as $p_{\text{old}}(\bm{z}|\bm{y})$. In E-Step, our goal is to sample $\bm{z}$ from $p_{\text{old}}(\bm{z}|\bm{y})$ using Langevin dynamics~\cite{welling2011bayesian}:
\begin{small}
\begin{equation}
    \bm{z}^{(\tau+1)} = \bm{z}^{(\tau)} + \frac{\delta^2}{2}\left[\frac{\partial}{\partial\bm{z}} \log p_{\text{old}}(\bm{z}|\bm{y}) \right] \bigg\vert_{\bm{z}=\bm{z}^{(\tau)}}
                                        + \delta \bm{\zeta}^{(\tau)},
    \label{eq:langevin}
\end{equation}
\end{small}
\hspace{-1mm}where 
$\tau$ indexs the time step for Langevin dynamics, $\delta$ denotes the step size, $\bm{\zeta}$ is the Gaussian white
noise used to prevent trapping into local modes. A key note to calculate Eq.~\eqref{eq:langevin} is
$\frac{\partial}{\partial\bm{z}} \log p_{\text{old}}(\bm{z}|\bm{y})=\frac{\partial}{\partial\bm{z}} \log p_{\text{old}}(\bm{z},\bm{y})$, and the detailed
calculation can be found in appendix.

In practice, a small trick to accelerate the convergence speed of Monte Carlo sampling in Eq.~\eqref{eq:langevin} is to start
from the previous updated $\bm{z}$ in each learning iteration. We empirically found that it performs very stably and well
by simply sampling 10 times according to Eq.~\eqref{eq:langevin}.

\begin{algorithm}[tb]
    \caption{Inference procedure for the proposed method}
    \label{alg:EM_algotithm}
    \begin{algorithmic}[1]
        \REQUIRE observed LR image, hyper-paramter settings.
        \ENSURE the super-resolved HR image $I^{\text{HR}}$. \\
        \STATE Initialize the model parameters $\{\bm{\alpha}, \bm{L}, \bm{\lambda}\}$ and the latent variable $\bm{z}$. \\
        \WHILE {not converged}
            \STATE \textbf{E-Step:} Sample the latent variable $\bm{z}$ from $p_{\text{old}}(\bm{z}|\bm{y})$ following
                                    Eq.~\eqref{eq:langevin}. \\
            \STATE \textbf{M-Step:} (a) Update parameters $\bm{\alpha}$ and $\bm{L}$ with fixed \\
                                        \hspace{1.85cm}$\bm{\lambda}$ according to Eq.~\eqref{eq:updata_alpha_xi}. \\
            \STATE \hspace{1.25cm}  (b) Update noise variance parameter $\bm{\lambda}$ with \\
                                        \hspace{1.85cm}fixed $\bm{\alpha}$ and $\bm{L}$ according to
                                        Eq.~\eqref{eq:upata_variance}. \\
        \ENDWHILE
        \STATE $I^{\text{HR}} = G(\bm{z};\bm{\alpha})$.
    \end{algorithmic}
\end{algorithm} 
\noindent\textbf{M-Step}.
Let's denote the sampled latent variable in E-step as $\tilde{\bm{z}}$, M-Step aims to maximize the approximate lower
bound of Eq.~\eqref{eq:log_map_goal} w.r.t. the model parameters $\{\bm{\alpha}, \bm{L}, \bm{\lambda}\}$:  
\begin{small}
\begin{align}
        \max_{\bm{\alpha}, \bm{L}, \bm{\lambda}} Q(\bm{\alpha}, \bm{L}, \bm{\lambda}) 
        &= \int p_{\text{old}}(\bm{z}|\bm{y})\log p(\bm{y}|\bm{\alpha},\bm{L}, \bm{\lambda}, \bm{z})
                 p(\bm{\alpha}|\bm{z}) p(\bm{z}) \mathrm{d}\bm{z} \notag \\ 
        &\approx \log p(\bm{y}|\bm{\alpha},\bm{L},\bm{\lambda},\tilde{\bm{z}})p(\bm{\alpha}|\tilde{\bm{z}})p(\tilde{\bm{z}}).
   \label{eq:Q_function}
\end{align}
\end{small}
\hspace{-1mm}Equivalently, Eq.~\eqref{eq:Q_function} can be reformulated into a minimization problem as follows:
\begin{small}
\begin{align}
    \min_{\bm{\alpha}, \bm{L}, \bm{\lambda}} E(\bm{\alpha}, \bm{L}, \bm{\lambda}) 
    &= \frac{1}{2} \left\Vert \frac{1}{\bm{\lambda}} \odot \Big\{\bm{y} -
                \big[ G(\tilde{\bm{z}};\bm{\alpha}) * h(\bm{L})\big] \downarrow_s^d \Big\} \right\Vert_2^2 \notag \\
    &\hspace{0.4cm}+\rho\sum_{k=1}^2 \left\vert f_k * G(\tilde{\bm{z}};\bm{\alpha}) \right\vert^{\gamma}.
    \label{eq:Q_fun_minimization}
\end{align}
\end{small}
\hspace{-1mm}To solve Eq.~\eqref{eq:Q_fun_minimization}, we alternately update the model parameters
$\{\bm{\alpha}$, $\bm{L}\}$ and $\bm{\lambda}$. Specifically, for $\bm{\alpha}$ and $\bm{L}$, they can be directly optimized by SGD based on the back-propagation (BP)
algorithm~\cite{rumelhart1986learning}:
\begin{small}
\begin{equation}
    \bm{W}_{\text{new}} = \bm{W}_{\text{old}} - \eta \frac{\partial}{\partial\bm{W}}E(\bm{\alpha}, \bm{L}, \bm{\lambda}),
                          ~ ~ \bm{W} \in \{\bm{\alpha}, \bm{L}\},
    \label{eq:updata_alpha_xi}
\end{equation}
\end{small}
\hspace{-1mm}where $\eta$ is the learning rate.
Actually, we adopt the more advanced Adam~\cite{Kingma2015} algorithm to update $\bm{\alpha}$ and $\bm{L}$ instead of the
SGD strategy of Eq.~\eqref{eq:updata_alpha_xi}, which empirically makes it converge much faster.

For the noise variance $\bm{\lambda}$, we consider $\lambda_i$ in the $p \times p$ patch centered at
the $i$-th pixel. Fortunately, based on the i.i.d. Gaussian assumption within this image patch, we have
the following closed-form solution for $\lambda_i$:
\begin{small}
\begin{equation}
    \lambda_i = \frac{1}{p^2}\sum_{j \in N(i)}
                 \left\{ y_j - \Big[ \big(
                        G(\tilde{\bm{z}};\bm{\alpha}_{\text{old}}) * h(\bm{L}_{\text{old}})
                                        \big) \downarrow_s^d \Big]_j \right\}^2,
    \label{eq:upata_variance}
\end{equation}
\end{small}
\hspace{-1mm}where $N(i)$ is the index set of the pixels in the $p \times p$ patch centered at $i$.

It should be noted that the first term of \eqref{eq:Q_fun_minimization} can be regarded as
a re-weighted $L_2$ loss with weight $\frac{1}{\bm{\lambda}}$, which is automatically updated through Eq.~\eqref{eq:upata_variance} during 
optimization. Detailed description of the proposed EM algorithm is presented in Algorithm~\ref{alg:EM_algotithm}.

\begin{table*}[t]
    \centering
    \caption{\footnotesize Averaged PSNR/SSIM/LPIPS results of the comparison methods under different degraded combinations on Set14. The best results are highlighted in \textbf{bold}.
    The \textcolor[gray]{0.5}{gray} results indicate unfair comparisons due to the mismatched degradations. Note that the results are averaged on six degradations with different blur
    kernels as shown in Fig.~\ref{fig:kernels_six} on Set14.}
    \label{tab:syn_set14}
    \vspace{-3mm}
    \scriptsize
    \begin{tabular}{@{}C{1.6cm}@{}|@{}C{2.2cm}@{}|@{}C{1.1cm}@{}|@{}C{1.75cm}@{}|@{}C{1.75cm}@{}|
                                                                 @{}C{1.75cm}@{}|@{}C{1.75cm}@{}|
                                                                 @{}C{1.75cm}@{}|@{}C{1.75cm}@{}|@{}C{1.85cm}@{}}
        \Xhline{0.8pt}
        \multirow{2}*{\makecell{Noise\\types}} & \multirow{2}*{Scale} & \multirow{2}*{Metrics} & \multicolumn{7}{c}{Methods} \\
        \Xcline{4-10}{0.4pt}
        &   &   & CSC~\cite{gu2015convolutional} &RCAN~\cite{zhang2018image} &ZSSR-B~\cite{shocher2018zero} &ZSSR-NB~\cite{shocher2018zero} &DoubleDIP~\cite{ren2020neural} &DIPFKP~\cite{liang2021flow} & BSRDM (ours)  \\
        \hline \hline
        \multirow{9}*{Case 1}  & \multirow{3}*{$\times$2}   & PSNR$\uparrow$     & 24.87    & \textcolor[gray]{0.5}{24.99}   & \textcolor[gray]{0.5}{25.04}   & \textbf{30.27}  & 23.98    & 27.45    & 29.56  \\
                               &                            & SSIM$\uparrow$     & 0.686    & \textcolor[gray]{0.5}{0.690}   & \textcolor[gray]{0.5}{0.701}   & \textbf{0.841}  & 0.637    & 0.752    & 0.815  \\
                               &                            & LPIPS$\downarrow$  & 0.318    & \textcolor[gray]{0.5}{0.321}   & \textcolor[gray]{0.5}{0.311}   & \textbf{0.263}  & 0.397    & 0.340    & 0.278  \\
        \Xcline{2-10}{0.4pt}
                               & \multirow{3}*{$\times$3}   & PSNR$\uparrow$     & 21.96    & \textcolor[gray]{0.5}{22.02}   & \textcolor[gray]{0.5}{22.06}   & 26.49           & 20.38    & 26.59    & \textbf{28.19}  \\
                               &                            & SSIM$\uparrow$     & 0.551    & \textcolor[gray]{0.5}{0.553}   & \textcolor[gray]{0.5}{0.566}   & 0.741           & 0.498    & 0.712    & \textbf{0.768}  \\
                               &                            & LPIPS$\downarrow$  & 0.397    & \textcolor[gray]{0.5}{0.390}   & \textcolor[gray]{0.5}{0.391}   & 0.362           & 0.469    & 0.383    & \textbf{0.328}  \\
        \Xcline{2-10}{0.4pt}
                               & \multirow{3}*{$\times$4}   & PSNR$\uparrow$     & 20.18    & \textcolor[gray]{0.5}{20.08}   & \textcolor[gray]{0.5}{20.23}   & 23.73           & 17.98    & 25.66    & \textbf{26.76}  \\
                               &                            & SSIM$\uparrow$     & 0.475    & \textcolor[gray]{0.5}{0.474}   & \textcolor[gray]{0.5}{0.490}   & 0.618           & 0.394    & 0.679    & \textbf{0.720}  \\
                               &                            & LPIPS$\downarrow$  & 0.464    & \textcolor[gray]{0.5}{0.452}   & \textcolor[gray]{0.5}{0.460}   & 0.522           & 0.533    & 0.419    & \textbf{0.381}  \\
        \hline \hline
        \multirow{9}*{Case 2}  & \multirow{3}*{$\times$2}   & PSNR$\uparrow$     & 24.43    & \textcolor[gray]{0.5}{24.52}   & \textcolor[gray]{0.5}{24.72}   & 26.73           & 23.42    & 26.95    & \textbf{28.01}  \\
                               &                            & SSIM$\uparrow$     & 0.648    & \textcolor[gray]{0.5}{0.651}   & \textcolor[gray]{0.5}{0.671}   & 0.723           & 0.618    & 0.734    & \textbf{0.771}  \\
                               &                            & LPIPS$\downarrow$  & 0.404    & \textcolor[gray]{0.5}{0.404}   & \textcolor[gray]{0.5}{0.385}   & 0.387           & 0.427    & 0.385    & \textbf{0.359}  \\
        \Xcline{2-10}{0.4pt}
                               & \multirow{3}*{$\times$3}   & PSNR$\uparrow$     & 21.70    & \textcolor[gray]{0.5}{21.73}   & \textcolor[gray]{0.5}{21.81}   & 24.74           & 20.03    & 25.31    & \textbf{26.24}  \\
                               &                            & SSIM$\uparrow$     & 0.523    & \textcolor[gray]{0.5}{0.526}   & \textcolor[gray]{0.5}{0.544}   & 0.657           & 0.475    & 0.662    & \textbf{0.706}  \\
                               &                            & LPIPS$\downarrow$  & 0.493    & \textcolor[gray]{0.5}{0.495}   & \textcolor[gray]{0.5}{0.481}   & 0.469           & 0.516    & 0.468    & \textbf{0.443}  \\
        \Xcline{2-10}{0.4pt}
                               & \multirow{3}*{$\times$4}   & PSNR$\uparrow$     & 20.03    & \textcolor[gray]{0.5}{19.99}   & \textcolor[gray]{0.5}{19.98}   & 23.79           & 18.02    & 24.18    & \textbf{24.79}  \\
                               &                            & SSIM$\uparrow$     & 0.454    & \textcolor[gray]{0.5}{0.454}   & \textcolor[gray]{0.5}{0.475}   & 0.619           & 0.376    & 0.608    & \textbf{0.648}  \\
                               &                            & LPIPS$\downarrow$  & 0.553    & \textcolor[gray]{0.5}{0.556}   & \textcolor[gray]{0.5}{0.543}   & 0.521           & 0.586    & 0.529    & \textbf{0.507}  \\
        \Xhline{0.8pt}
    \end{tabular}
    \vspace{-6mm}
\end{table*}

\section{Experimental Results}\label{sec:experiments}
We conducted extensive experiments to verify the effectiveness of the proposed method in this section.
For ease of presentation, we briefly denote our \textbf{b}lind \textbf{s}uper-\textbf{r}esolution method with elaborate
\textbf{d}egradation \textbf{m}odeling on noise and kernel as BSRDM in the rest of this paper.

\subsection{Experimental Setup}
\noindent\textbf{Model Settings}. Throughout the experiments, we empirically set the hyper-paramters $\rho$ and $\gamma$ to
be $0.2$ and $\sfrac{2}{3}$, respectively. The setting on $\gamma$ lies on the fact that the hyper-Laplacian with
exponent $\gamma=\sfrac{2}{3}$ is a better model of image gradient than a Laplacian or Gaussian~\cite{krishnan2009fast}.
To update the model parameters $\bm{\alpha}$ and $\bm{L}$ in M-Step, the Adam~\cite{Kingma2015} algorithm with default
settings in Pytorch~\cite{paszke2019pytorch} is used. The learning rates for $\bm{\alpha}$ and $\bm{L}$ are set as
$2e\text{-}3$ and $5e\text{-}3$, respectively. As for the patch size $p$ of the noise model, we provide two different
settings. For the synthetic Gaussian noise in Sec.~\ref{subsec:syn_experiments}, we regard the whole image as one special
image patch. While for synthetic camera sensor noise in Sec.~\ref{subsec:syn_experiments} and
real image noise in Sec.~\ref{subsec:real_experiments}, we set $p$ to 15. For fair comparison, the quantitative results
of our method are averaged by running it five times with different random seeds.

\noindent\textbf{Comparison Methods}. To evaluate BSRDM, we compare it against five methods, including one learning-based
method RCAN~\cite{zhang2018image}, and four model-based methods, namely CSC~\cite{gu2015convolutional},
ZSSR~\cite{shocher2018zero}, DoubleDIP~\cite{ren2020neural}, and DIPFKP~\cite{liang2021flow}. Specifically, RCAN is
a blind SISR method trained under the bicubic degradation; CSC attempts to recover high frequency image details using
convolutional sparse coding; ZSSR is a zero-shot method that exploits the patch recurrence across scales in a single image;
DoubleDIP and DIPFKP are both blind SISR methods but with different kernel priors. In the synthetic experiments
of Sec.~\ref{subsec:syn_experiments}, we consider both blind and non-blind settings for ZSSR, and denote them
as ``ZSSR-B'' and ``ZSSR-NB'', respectively. For ZSSR-B, we use the default setting of its official code, in which the
degradation model is assumed to be a bicubic downsampler followed by AWGN noise. While for ZSSR-NB, the ground truth
blur kernels are pre-provided by us. Noted that ZSSR, DoubleDIP, DIPFKP, and BSRDM all employ deep CNN to generate HR image.
Thus the comparison with them can better verify the marginal effects brought up by the noise and kernel modeling in BSRDM.
\begin{figure}[t]
    \centering
    \includegraphics[scale=0.64]{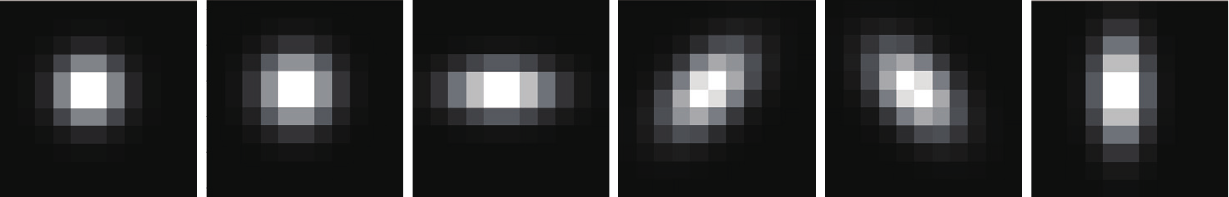}
    \vspace{-2mm}
    \caption{Six Gaussian kernels used to synthesize the LR images.}
    \label{fig:kernels_six}
    \vspace{-3mm}
\end{figure}


\subsection{Evaluation on Synthetic Data}\label{subsec:syn_experiments}
\begin{figure}[t]
    \centering
    \includegraphics[scale=0.89]{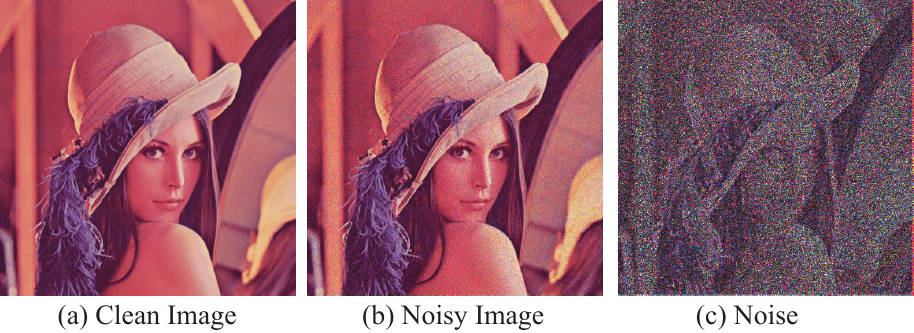}
    \vspace{-7mm}
    \caption{\footnotesize Illustration of camera sensor noise of Case 2. From left to right: (a) clean image; (b) simulated noisy image with camera sensor noise;
        (c) absolute residual (or noise) between (a) and (b).}
    \label{fig:noise}
    \vspace{-4mm}
\end{figure}

In this part, we quantitatively evaluate different methods on two commonly-used datasets, i.e., Set14~\cite{zeyde2010single} and DIV2K100~\cite{agustsson2017ntire}.
DIV2K100 contains 100 high resolution images of the validation set of DIV2K, and we crop a $1024 \times 1024$ patch around the center from each image in our experiments due
to GPU memory limitation. The LR images are synthesized via Eq.~\eqref{eq:degradation_generation}. To conduct a thorough comparison, we consider diverse degradations combined with different
blur kernels and noise types. For blur kernels, two isotropic Gaussian kernels with different widths (i.e., $1.2$ and $2.0$) and four anisotropic Gaussian kernels are
chosen as shown in Fig.~\ref{fig:kernels_six}.
Furthermore, we consider two noise types as follows:
\begin{myitemize}
\item Case 1: Gaussian noise with noise level $2.55$, which is widely used in current SISR literatures~\cite{zhang2020deep,yue2019variational}.
    \item Case 2: Camera sensor noise simulated by~\cite{brooks2019unprocessing,guo2019toward}, one typical example is shown in Fig.~\ref{fig:noise}.
\end{myitemize}
\begin{figure*}[t]
    \centering
    \includegraphics[scale=0.81]{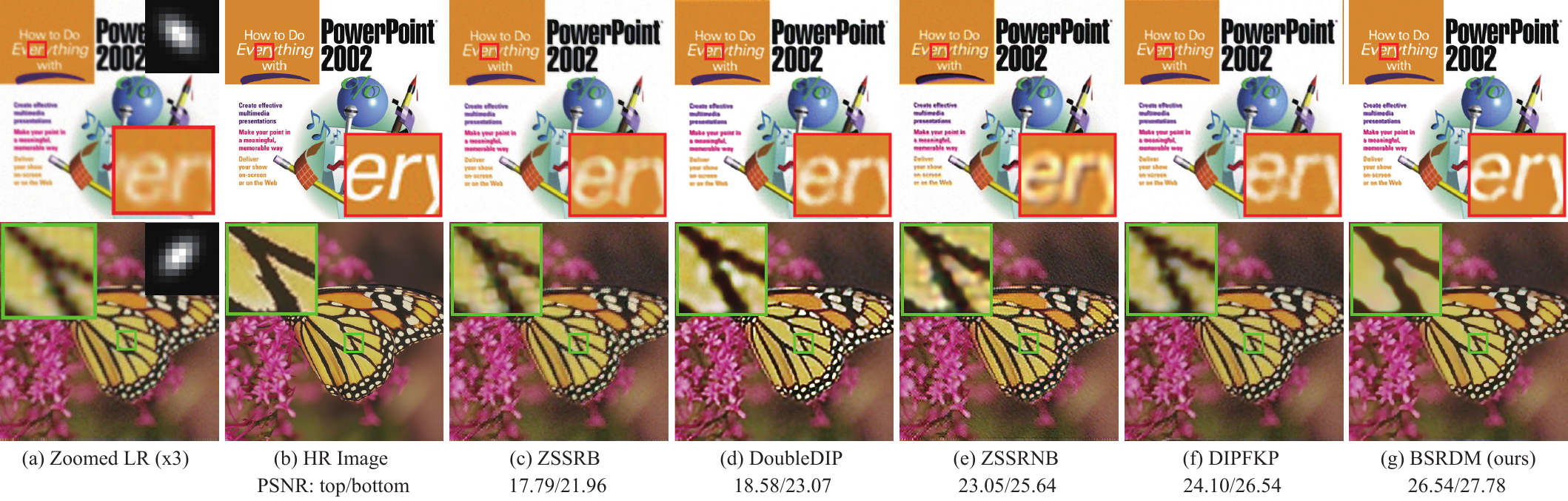}
    \vspace{-8mm}
    \caption{\footnotesize Super-resolution results of different methods for two degradations with Gaussian noise (top row) and camera sensor noise (bottom row)
            under scale factor 3 on Set14. The blur kernel is shown on the upper-right conner of the zoomed LR image.}
    \label{fig:syn_x3}
    \vspace{-2mm}
\end{figure*}
\begin{figure*}[t]
    \centering
    \includegraphics[scale=0.81]{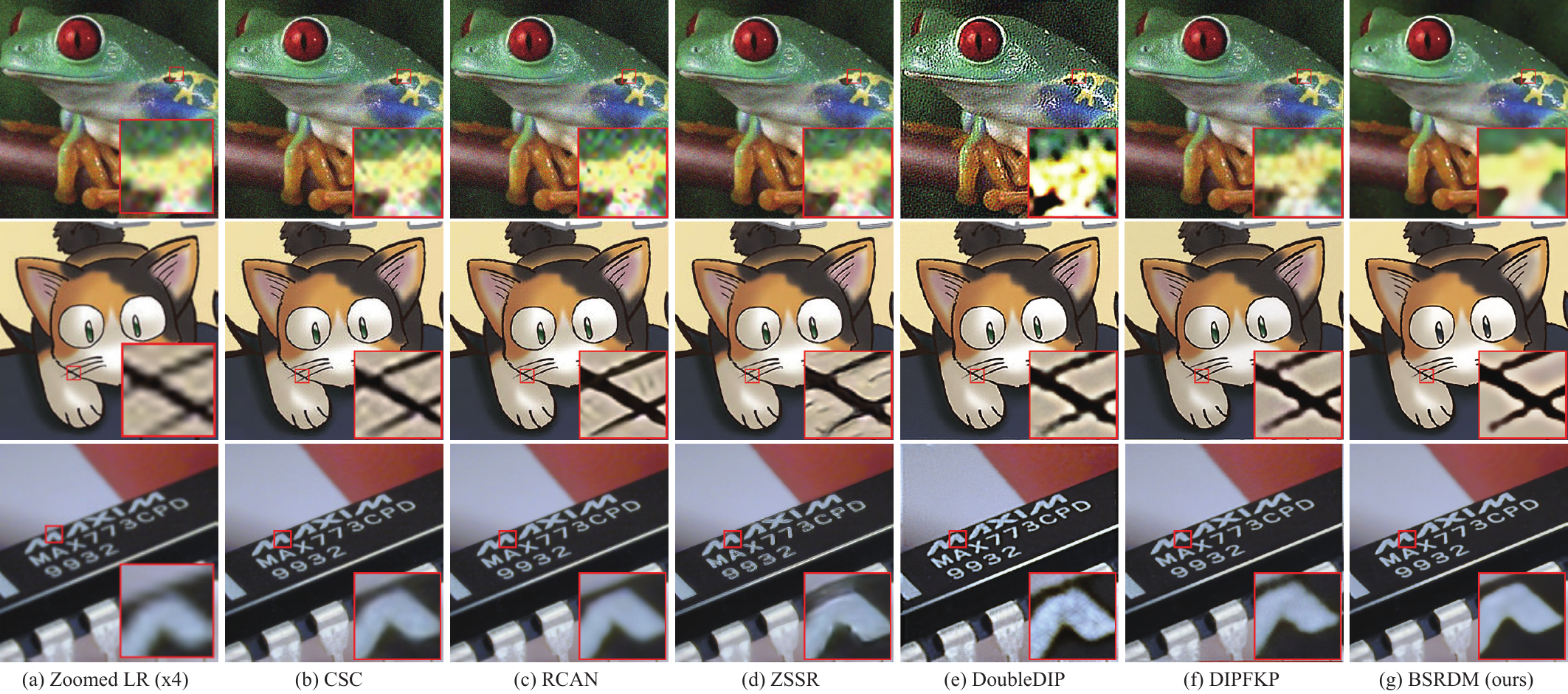}
    \vspace{-7mm}
    \caption{\footnotesize Three super-resolution results of different methods on real LR images with scale factor 4. Please zoom in for best view.}
    \label{fig:real_x4}
    \vspace{-2mm}
\end{figure*}
Especially, the noise in Case 2 is very close to real camera noise, and is thus suitable for evaluating different methods under the degradations with complicated real noise.
As for the quantitative metrics, except for the commonly-used PSNR and SSIM~\cite{wang2004image}, we also adopted LPIPS~\cite{zhang2018unreasonable} to compare the perceptual similarity between
the recovered HR image and ground-truth. Note that PSNR and SSIM are calculated in the luminance channel like most of the SISR literatures, while LPIPS is directly calculated in RGB channels.

\noindent\textbf{Comparison with SotA Methods}. Table~\ref{tab:syn_set14} lists the PSNR, SSIM, and LPIPS results
of different methods under diverse degradations on Set14. The comparison on DIV2K100 can be found in appendix. From
Table~\ref{tab:syn_set14}, we can see that the proposed BSRDM achieves the best or at least the second best results
for all degradations. For the degradation with scale factor 2 and Gaussian noise, ZSSR-NB achieves the best performance.
While for the degradation with scale factor 2 and camera sensor noise, BSRDM outperforms ZSSR-NB, indicating that BSRDM
is able to handle more complicated noise due to its non-i.i.d. noise modeling. Comparing with current state-of-the-arts
(SotA) method DIPFKP, the evidently superiority of BSRDM demonstrates the importance of the noise and kernel
modeling in SISR, since they both use the same network architecture to generate HR image even though BSRDM has fewer
parameters (see Sec.~\ref{subsec:speed}).

Two visual results on Gaussian noise (top row) and camera sensor noise (bottom row) are shown in Fig.~\ref{fig:syn_x3}. Note that we only display the five best methods
due to page limitation, and the complete results can be found in appendix. We can easily observe that: 1) In the case of Gaussian noise, all the comparison methods can remove
such simple AWGN noise. Due to the better kernel modeling, the proposed BSRDM evidently achieves sharper results. 2) Under camera sensor noise, the recovered images
of the four comparison methods still contain some obvious noises or artifacts, mainly because their i.i.d. Gaussian noise assumption largely deviates from the true noise distribution.
On the contrary, BSRDM is able to remove most of the noises and preserves clear image details. This demonstrates the effectiveness of the proposed non-i.i.d. noise assumption
under the complicated noise.

\noindent\textbf{Ablation Studies}. The core contributions of this paper mainly include the non-i.i.d. noise modeling manner and the constructed kernel prior EKP.
To justify their effectiveness, we design two baseline methods. In the first baseline (denoted as \textit{Baseline1}), we replace the non-i.i.d. noise
assumption with the conventional i.i.d. one. Similarly, in the second baseline (denoted as \textit{Baseline2}), the proposed EKP kernel prior is replaced with
FKP~\cite{liang2021flow}, which is the current most effective kernel prior to our best knowledge.

We compare BSRDM with these two baselines on different degradations that combine the six blur kernels in Fig.~\ref{fig:kernels_six} and camera sensor noise under scale factor 2 on Set14.
The detailed results are listed in Table~\ref{tab:ablation}. Firstly, comparing with \textit{Baseline1},
the performance gain of BSRDM is mainly brought up by the non-i.i.d. noise assumption, which makes it be able to better deal with such signal-depedent camera sensor noise. Secondly,
the superiority of BSRDM over \textit{Baseline2} indicates that our proposed kernel prior EKP is more effective than FKP as analysed in Sec.~\ref{subsec:model}.

\begin{figure}[t]
    \centering
    \includegraphics[scale=0.59]{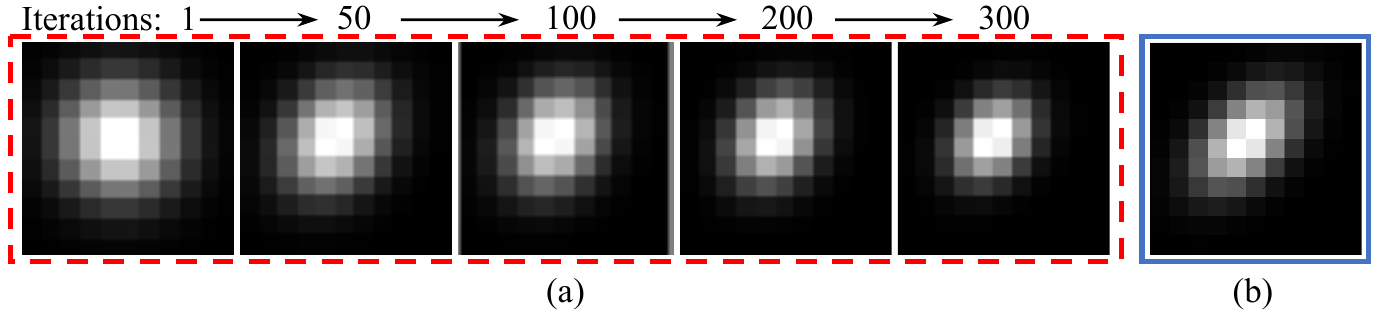}
    \vspace{-8mm}
    \caption{(a) Estimated kernels by our method at the 1st, 50th, 100th, 200th, and 300th iterations, (b) Ground-truth blur kernel.}
    \label{fig:kernel_pre}
    \vspace{-3mm}
\end{figure}

Figure~\ref{fig:kernel_pre} displays the estimated blur kernels by our method in different iterations during optimization. Note that the blur kernel is initialized as an isotropic Gaussian kernel with
width $s$ (i.e., the 1st iteration), where $s$ is the scale factor. From this figure, we can see that the kernel is gradually adjusted toward the ground truth. After 300
iterations, the estimated kernel is very close to the ground truth, which facilitates a good super-resolution result.

\begin{table}[t]
    \centering
    \caption{Ablation studies under camera sensor noise with scale factor 2 on Set14. The PSNR/SSIM/LPIPS results are averaged on the six kernel settings as shown in Fig.~\ref{fig:kernels_six}.}
    \label{tab:ablation}
    \vspace{-3mm}
    \scriptsize
    \begin{tabular}{@{}C{1.60cm}@{}|@{}C{1.00cm}@{}|@{}C{1.15cm}@{}|
                                   @{}C{0.95cm}@{}|@{}C{0.95cm}@{}|@{}C{2.50cm}@{}}
        \Xhline{0.8pt}
        \multirow{2}*{Methods}        & \multicolumn{2}{c|}{Noise Assumption}  & \multicolumn{2}{c|}{Kernel Prior}   & \multirow{2}*{PSNR / SSIM / LPIPS} \\
        \Xcline{2-5}{0.4pt}
                                      & I.i.d          & Non-i.i.d.           & FKP           & EKP                  &                        \\
        \Xhline{0.4pt}
        \textit{Baseline1}            & $\checkmark$   &                      &               & $\checkmark$         & 27.76 / 0.768 / 0.373   \\
        \Xhline{0.4pt}
        \textit{Baseline2}            &                & $\checkmark$         & $\checkmark$  &                      & 27.52 / 0.765 / 0.362   \\
        \Xhline{0.4pt}
        BSRDM (ours)                  &                & $\checkmark$         &               & $\checkmark$         & 28.01 / 0.771 / 0.359   \\
        \Xhline{0.8pt}
    \end{tabular}
    \vspace{-3mm}
\end{table}
\subsection{Evaluation on Real Data}\label{subsec:real_experiments}
To further justify the effectiveness of BSRDM in real SISR task, we evaluate it on RealSRSet~\cite{zhang2021designing}, which contains 20 real images from internet or the existing testing
datasets~\cite{ignatov2017dslr,martin2001database,matsui2017sketch,zhang2018ffdnet}. Figure~\ref{fig:real_x4} shows three typical examples that include different scenarios in SISR, i.e., natural
image (top row), cartoon image (middle row), and text image (bottom row).
It can be easily seen that the proposed BSRDM achieves evidently better visual results than the other comparison methods.
In the first and second examples (bottom and middle row of Fig.~\ref{fig:real_x4}), the LR images contain some obvious camera sensor noises or
artifacts. Most of the comparison methods cannot finely deal with these cases, and tend to enlarge the noises or artifacts after super-resolution. The proposed BSRDM is able to remove most of these noises or
artifacts and preserve clearer image structures due to its more robust non-i.i.d. noise modeling. For the commonly-used ``chip'' example (bottom row of Fig.~\ref{fig:real_x4}) in SISR,
the super-resolution results of the comparison methods are all very blurry, which may be caused by the fact that the estimated kernel does not match with the true one. On the contrary, BSRDM can
obtain a relatively sharper and cleaner HR image because the proposed EKP makes it easier to estimate a rational blur kernel.

As pointed out by~\cite{zhang2021designing}, we also find that current non-reference metrics (e.g., NIQE~\cite{mittal2012making}, NRQM~\cite{Ma2017}, and PI~\cite{agustsson2017ntire}) are not
consistent with our perceptual visual system in real SISR task. We put the detailed quantitative comparisons in terms of non-reference metrics and more visual results in appendix due to page limitation.

\begin{table}[t]
    \centering
    \caption{\footnotesize Comparison results of different methods on model size (K) and runtime (s).}
    \label{tab:capacity_time}
    \vspace{-3mm}
    \scriptsize
    \begin{tabular}{@{}C{1.8cm}@{}|@{}C{0.55cm}@{}@{}C{0.55cm}@{}@{}C{0.55cm}@{}|
                                   @{}C{0.55cm}@{}@{}C{0.55cm}@{}@{}C{0.55cm}@{}|
                                   @{}C{0.55cm}@{}@{}C{0.55cm}@{}@{}C{0.55cm}@{}|
                                   @{}C{0.55cm}@{}@{}C{0.55cm}@{}@{}C{0.55cm}@{}}
        \Xhline{1.0pt}
        Methods        & \multicolumn{3}{c|}{ZSSR}     & \multicolumn{3}{c|}{DoubleDIP}
                       & \multicolumn{3}{c|}{DIPFKP}   & \multicolumn{3}{c}{BSRDM} \\
        \Xhline{0.4pt}
        Scale          & $\times$2  & $\times$3  & $\times$4  & $\times$2  & $\times$3  & $\times$4
                       & $\times$2  & $\times$3  & $\times$4  & $\times$2  & $\times$3  & $\times$4 \\
        \Xhline{0.4pt}
        Time (s)       & 56         &  117       &  235       &  90        &  194       &  361
                       & 91         &  190       &  333       &  53        &  108       &  190 \\ 
        \Xhline{0.4pt}
        \# parameters (K)   & \multicolumn{3}{c|}{225}        &  \multicolumn{3}{c|}{2396}          
                            & \multicolumn{3}{c|}{2396}       &  \multicolumn{3}{c}{762} \\
        \Xhline{1.0pt}
    \end{tabular}
    \vspace{-3mm}
\end{table}
\subsection{Comparison on Model Size and Runtime} \label{subsec:speed}
Table~\ref{tab:capacity_time} lists the comparison results on model size (number of parameters) and runtime with existing model-based SISR methods.
For fair comparison, we consider three typical methods (i.e., ZSSR, DoubleDIP, and DIPFKP) that are all accelerated by GPU, and the runtime results
in Table~\ref{tab:capacity_time} are tested on a GeForce RTX 2080 Ti GPU. Specifically, we fix the LR image size as $256 \times 256$
and count the elapsed time of super-resolving it to size of $512 \times 512$, $768 \times 768$, and $1024 \times 1024$ with scale factor $2$, $3$, and $4$,
respectively. From Table~\ref{tab:capacity_time}, it can be easily observed that: 1) Our BSRDM has a moderate number of parameters comparing with other methods.
2) Even though BSRDM contains more parameters than ZSSR, it still has the similar speed with ZSSR. What's more, BSRDM is a little faster than ZSSR under
scale factor 4. 3) Comparing with current SotA method DIPFKP, BSRDM is not only with faster speed but also much fewer parameters.
Taking all of the comparisons on model size, runtime, and the performances on SISR into consideration, it should be rational to say that BSRDM
is effective and potentially useful in real applications.

\section{Conclusion}
In this paper, we have proposed a new blind SISR method under the probabilistic framework, which elaborately considers the
degradation modeling on noise and kernel. Specifically, to better fit the complicated real noise, a patch-based non-i.i.d.
noise distribution is adopted in our method. As for the blur kernel, we construct an explicit yet effective kernel prior
named EKP and apply it in the proposed method. Through extensive experiments, we have verifed the effectiveness and
superiority of the proposed method on synthetic and real datasets. We believe that this work can benefit the blind SISR
research community.

\noindent\textbf{Acknowledgement}. This work was partially supported by the National Key R\&D Program of
China (2020YFA0713900), the Hong Kong RGC GRF grant (project\# 17203119), the Macao Science and Technology Development
Fund under Grant 061/2020/A2, the China NSFC projects under contracts 61721002 and U1811461.

{\small
\bibliographystyle{ieee_fullname}
\bibliography{bsrdm_refer}
}

\newpage
\appendix
\section{Calculation Details on the E-Step}
Given current model parameters $\{\bm{\alpha}_{\text{old}}, \bm{L}_{\text{old}}, \bm{\lambda}_{\text{old}}\}$, we denote the posterior of $\bm{z}$ under them as $p_{\text{old}}(\bm{z}|\bm{y})$.
In E-Step, our goal is to sample $\bm{z}$ from $p_{\text{old}}(\bm{z}|\bm{y})$ using Langevin dynamics~\cite{welling2011bayesian}:
\begin{small}
\begin{align}
    \bm{z}^{(\tau+1)} & = \bm{z}^{(\tau)} + \frac{\delta^2}{2}\left[\frac{\partial}{\partial\bm{z}} \log p_{\text{old}}(\bm{z}|\bm{y}) \right] \bigg\vert_{\bm{z}=\bm{z}^{(\tau)}}
                                          + \delta \bm{\zeta}^{(\tau)} \notag \\
                      & = \bm{z}^{(\tau)} + \frac{\delta^2}{2}\left[\frac{\partial}{\partial\bm{z}} \log p_{\text{old}}(\bm{z},\bm{y}) \right] \bigg\vert_{\bm{z}=\bm{z}^{(\tau)}}
                                          + \delta \bm{\zeta}^{(\tau)} \notag \\
                      & = \bm{z}^{(\tau)} - \frac{\delta^2}{2}\left[\frac{\partial}{\partial\bm{z}}  g(\bm{z}) \right] \bigg\vert_{\bm{z}=\bm{z}^{(\tau)}} + \delta \bm{\zeta}^{(\tau)},
    \label{eq:langevin_supp}
\end{align}
\end{small}
where 
\begin{small}
\begin{align}
g(\bm{z}) & = \frac{1}{2} \left\Vert \frac{1}{\bm{\lambda}_{\text{old}}} \odot \Big\{ \bm{y} -  \big[G(\bm{z};\bm{\alpha}_{\text{old}}) * h(\bm{L}_{\text{old}})\big] \downarrow_s^d \Big\} \right\Vert_2^2   \notag \\
          & + \rho\sum_{k=1}^2 \left\vert f_k * G(\bm{z};\bm{\alpha}_{\text{old}}) \right\vert^{\gamma} + \frac{1}{2} \Vert \bm{z} \Vert^2_2,
    \label{eq:gz_langevin}
\end{align}
\end{small}
\hspace{-1mm}$\tau$ indexs the time step for Langevin dynamics, $\delta$ denotes the step size, $\bm{\zeta}$ is the Gaussian white noise used to
prevent trapping into local modes, $\odot$ represents the Hadamard product. As for the derivation of $g(\bm{z})$, we firstly   
factorize $p_{\text{old}}(\bm{z}, \bm{y})$ as follows:
\begin{equation}
    p_{\text{old}}(\bm{z}, \bm{y}) = p(\bm{y}|\bm{\alpha}_{\text{old}}, \bm{L}_{\text{old}}, \bm{\lambda}_{\text{old}}, \bm{z})
                                     p(\bm{\alpha}_{\text{old}}|\bm{z}) p(\bm{z}),
                     \label{eq:gz_calculation}
\end{equation}
where $p(\bm{y}|\bm{\alpha}_{\text{old}}, \bm{L}_{\text{old}}, \bm{\lambda}_{\text{old}}, \bm{z})$, $p(\bm{\alpha}_{\text{old}}|\bm{z})$, and $p(\bm{z})$ are
defined in Eq.~\eqref{eq:noise_likelihood}, Eq.~\eqref{eq:hyper_laplacian}, and Eq.~\eqref{eq:latent_prior}, respectively. By substituting these three terms into Eq.~\eqref{eq:gz_calculation}, we can
easily obtain the formulation in Eq.~\eqref{eq:gz_langevin} after simple derivation.

\section{Network Architecture}
\begin{figure*}[t]
    \centering
    \includegraphics[scale=0.75]{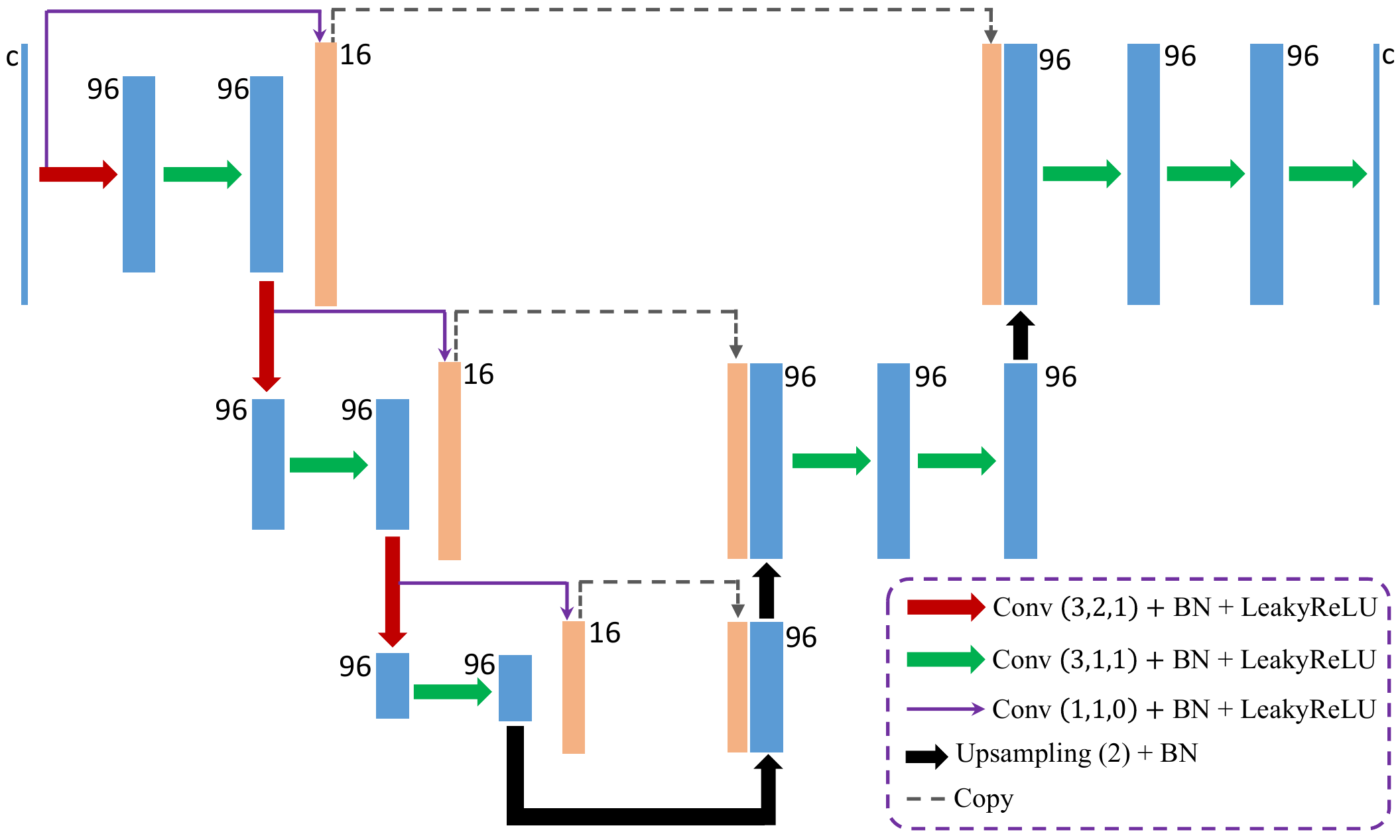}
    \vspace{-2mm}
    \caption{The detailed network architecture of the generator $G$. ``Conv ($k$,$p$,$s$)'' represents the 2-D convolution 
    operator with kernel size $k$, stride $s$ and reflection padding size $p$, ``BN'' represents the Batch Normlization layer,
    ``LeakyReLU'' represents the LeakyReLU activation function with negative slope $0.25$, and ``Upsampling ($s$)'' represents
    the nearest interpolation operator with scale factor $s$. The blue or orange rectangles denote the feature maps of the
    intermediate layers, and the numbers along them are the corresponding number of channels.}
    \label{fig:generator}
\end{figure*}

As for the generator $G$, we follow the ``hourglass'' archtechture in DIP~\cite{ulyanov2018deep}. However,
we used a very tiny version that contains much fewer parameters as shown in Sec.~\ref{subsec:speed}. The detailed network
architecture is shown in Fig.~\ref{fig:generator}. Note that, as for the upsampling operation, the nearest interpolation is employed.

\section{Experimental Results}
\begin{table}[t]
    \centering
    \caption{\footnotesize Performances of the proposed BSRDM with different settings of $\rho$ on Set14. The PSNR/SSIM/LPIPS results are all
    averaged on different degradations combined with camera sensor noise and six different blur kernels (see Fig.~\ref{fig:kernels_six}) under scale factor $2$.}
    \label{tab:hyper-rho}
    \vspace{-3mm}
    \scriptsize
    \begin{tabular}{@{}C{1.6cm}@{}|@{}C{2.20cm}@{}|@{}C{2.20cm}@{}|@{}C{2.20cm}@{}}
        \Xhline{0.8pt}
        \multirow{2}*{$\rho$} & \multicolumn{3}{c}{Metrics} \\
        \Xcline{2-4}{0.4pt}
                &{PSNR$\uparrow$}   &{SSIM$\uparrow$}   &{LPIPS$\downarrow$}  \\
        \Xhline{0.4pt}
        0      & 27.20    & 0.725   & 0.379 \\
        0.01   & 27.37    & 0.737   & 0.378 \\
        0.10   & 27.84    & 0.762   & 0.366 \\
        0.20   & 28.01    & 0.771   & 0.360 \\
        0.30   & 28.06    & 0.774   & 0.356 \\
        0.40   & 28.09    & 0.774   & 0.355 \\
        0.50   & 28.06    & 0.772   & 0.355 \\
        1.00   & 27.56    & 0.744   & 0.383 \\
        \Xhline{1.0pt}
    \end{tabular}
    \vspace{-3mm}
\end{table}
\begin{figure*}[t]
    \centering
    \includegraphics[scale=0.81]{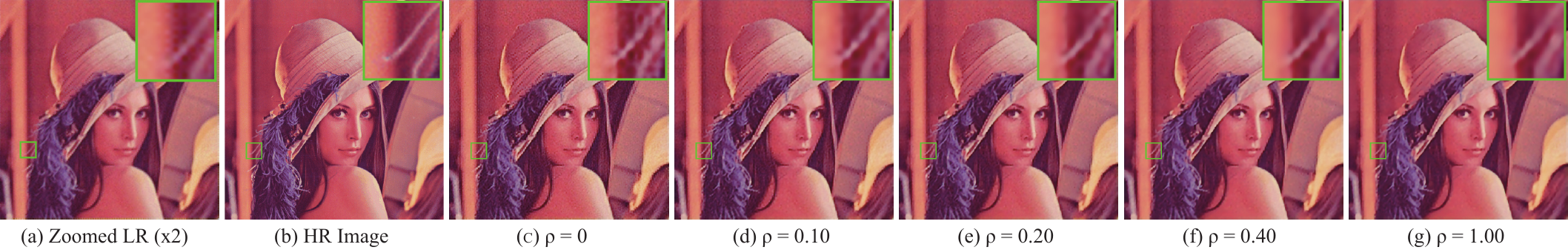}
    \vspace{-6mm}
    \caption{One typical example of the proposed method under different settings of $\rho$ for the degradation with camera sensor noise on Set14. From left to
    right: (a) the zoomed LR image, (b) the HR image, (c)-(d) the super-resolved results of BSRDM under different $\rho$ values.}
    \label{fig:rho_lenna}
\end{figure*}
\begin{table}[t]
    \centering
    \caption{\footnotesize Performances of the proposed BSRDM with different settings of $\gamma$ on Set14. The PSNR/SSIM/LPIPS results are all
    averaged on different degradations combined with camera sensor noise and six different blur kernels (see Fig.~\ref{fig:kernels_six}) under scale factor $2$.}
    \label{tab:hyper-gamma}
    \vspace{-3mm}
    \scriptsize
    \begin{tabular}{@{}C{1.6cm}@{}|@{}C{2.20cm}@{}|@{}C{2.20cm}@{}|@{}C{2.20cm}@{}}
        \Xhline{0.8pt}
        \multirow{2}*{$\gamma$} & \multicolumn{3}{c}{Metrics} \\
        \Xcline{2-4}{0.4pt}
                &{PSNR$\uparrow$}   &{SSIM$\uparrow$}   &{LPIPS$\downarrow$}  \\
        \Xhline{0.4pt}
        0.67    & 28.01   & 0.771   & 0.360 \\
        1.00    & 27.83   & 0.760   & 0.367 \\
        2.00    & 27.27   & 0.738   & 0.375 \\
        \Xhline{1.0pt}
    \end{tabular}
    \vspace{-2mm}
\end{table}
\begin{figure}[t]
    \centering
    \includegraphics[scale=0.91]{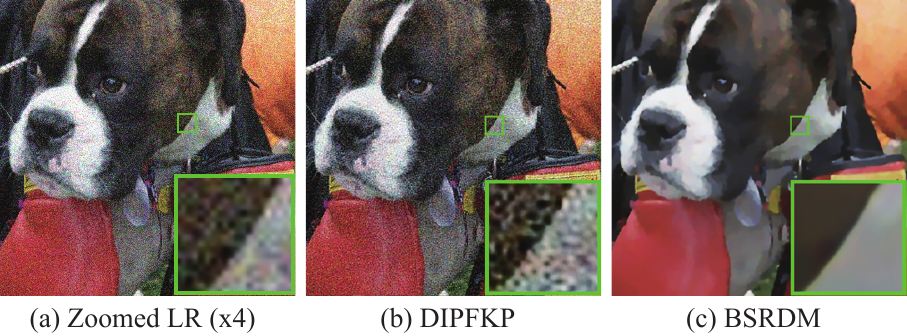}
    \vspace{-6mm}
    \caption{Visual super-resolution results of the ``dog'' example in RealSRSet~\cite{zhang2021designing}. From left to right: (a) the zoomed LR image, (b)-(c) the recovered HR images of 
             DIPFKP~\cite{liang2021flow} and the proposed BSRDM, respectively.}
    \label{fig:real_limit}
    \vspace{-3mm}
\end{figure}
\begin{figure*}[t]
    \centering
    \includegraphics[scale=0.81]{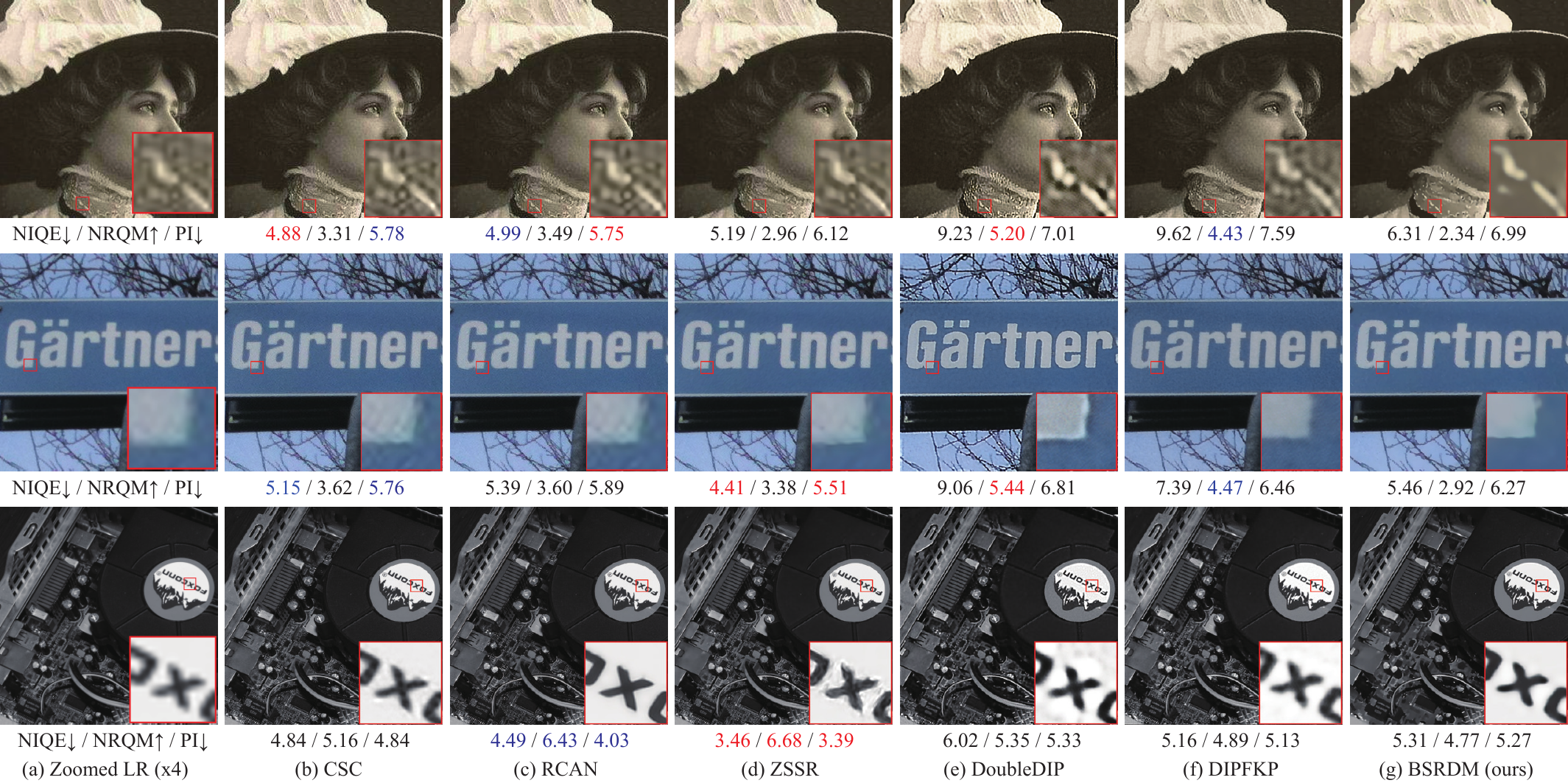}
    \vspace{-7mm}
    \caption{Three typical visual results on the RealSRSet~\cite{zhang2021designing} with scale factor 4. The best and second best non-reference metrics are highlighted in red 
             and blue. Please zoom in for best view.}
    \label{fig:real_x4_supp}
    \vspace{-3mm}
\end{figure*}
\begin{figure*}[t]
    \centering
    \includegraphics[scale=0.81]{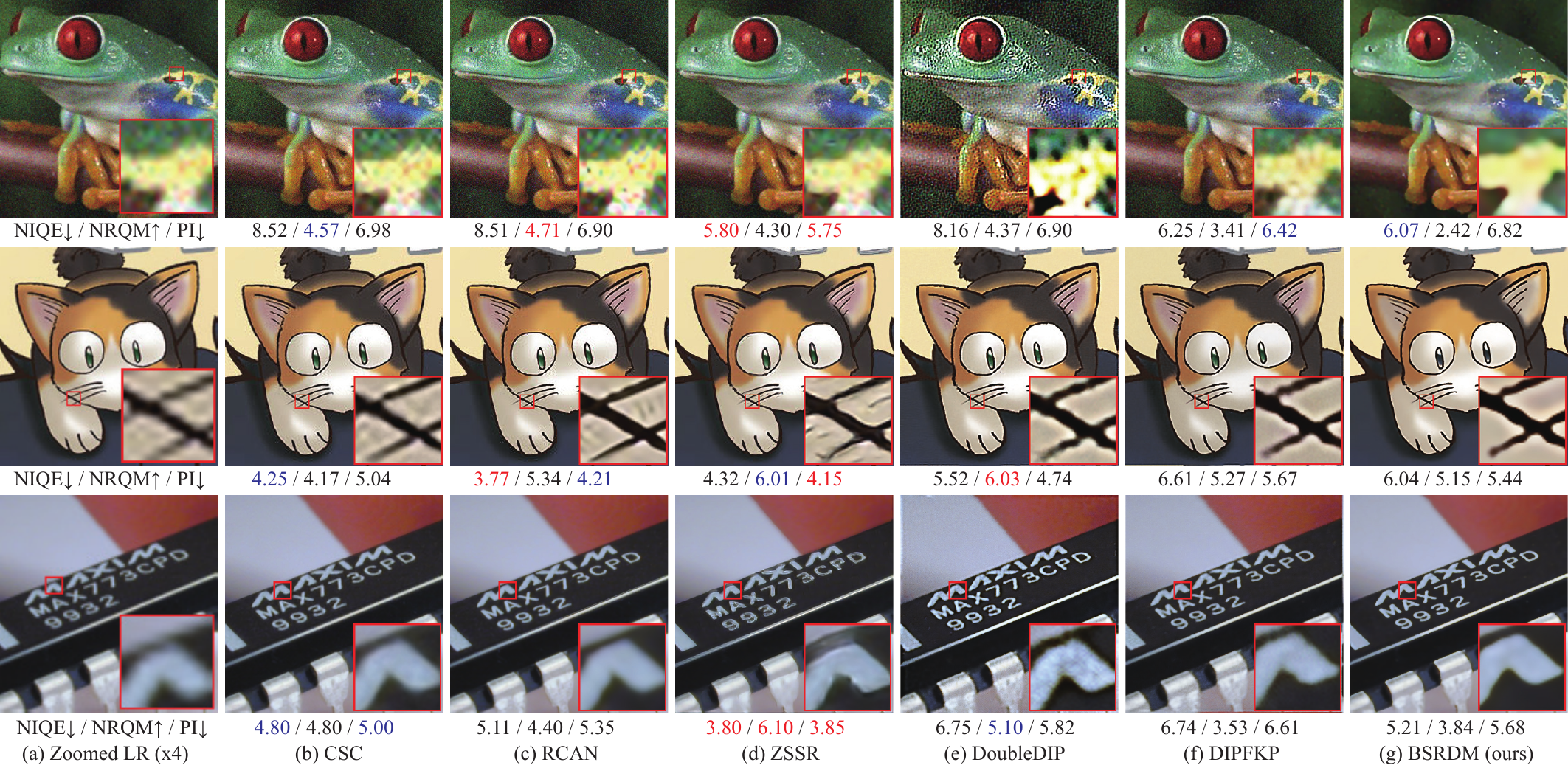}
    \vspace{-7mm}
    \caption{Three typical visual results on the RealSRSet~\cite{zhang2021designing} with scale factor 4. The best and second best non-reference metrics are highlighted in red 
        and blue. Note that this figure is the same with the Fig.~\ref{fig:real_x4}, but the non-reference metrics (i.e., NIQE, NRQM and PI) are
             additionally marked for each image. Please zoom in for best view.}
    \label{fig:real_x4_maintext}
\end{figure*}
\subsection{Hyper-parameter Analysis}
As shown in Sec.~\ref{subsec:model}, our proposed BSRDM mainly involves two hyper-parameters, i.e., $\rho$ and $\gamma$.
Next, we empirically analyse the sensitiveness of BSRDM to them.
\begin{table*}[t]
    \centering
    \caption{The non-reference NIQE, NRQM and PI comparison results of different methods on the RealSRSet data set. The best and second best results are highlighted in red and blue.}
    \label{tab:non-refer}
    \vspace{-3mm}
    \begin{tabular}{@{}C{2.2cm}@{}|@{}C{2.20cm}@{}|@{}C{2.20cm}@{}|
                                   @{}C{2.20cm}@{}|@{}C{3.00cm}@{}|
                                   @{}C{2.75cm}@{}|@{}C{2.75cm}@{}}
        \Xhline{1.0pt}
        \multirow{2}*{Metrics}  & \multicolumn{6}{c}{Methods}  \\
        \Xcline{2-7}{0.4pt}
                         & CSC~\cite{gu2015convolutional}  & RCAN~\cite{zhang2018image}  & ZSSR~\cite{shocher2018zero} & DoubleDIP~\cite{ren2020neural} & DIPFKP~\cite{liang2021flow}  & BSRDM (ours) \\
        \Xhline{0.4pt}
        NIQE$\downarrow$ & 5.87                            & \textcolor{blue}{5.61}      & \textcolor{red}{4.73}       & 7.29                           & 7.04                         & 6.23         \\
        NRQM$\uparrow$   & 4.16                            & 4.58                        & \textcolor{red}{5.36}       & \textcolor{blue}{5.22}         & 4.45                         & 3.99         \\
        PI$\downarrow$   & 5.85                            & \textcolor{blue}{5.51}      & \textcolor{red}{4.69}       & 6.04                           & 6.29                         & 6.12         \\
        \Xhline{1.0pt}
    \end{tabular}
    \vspace{-2mm}
\end{table*}

\textbf{Hyper-parameter $\rho$:} Intuitively, the hyper-paramter $\rho$ controls the relative importance of the hyper-Laplacian
prior in our method. Table \ref{tab:hyper-rho} lists the PSNR/SSIM performance of our proposed BSRDM under different $\rho$ values
on Set14~\cite{zeyde2010single}, and one corresponding visual results are shown in Fig.~\ref{fig:rho_lenna}. It can be easily seen that
BSRDM performs very stably and well in range of [0.2, 0.5], but larger $\rho$ value tends to produce more smooth results. Therefore, taking both of the
quantitative and qualitative results into consideration, we set $\rho$ to be $0.20$ in our experiments.

\textbf{Hyper-parameter $\gamma$:} The hyper-paramter $\gamma$ reflects the strength of the sparsity constraint on the image gradients.
The Eq.~\eqref{eq:hyper_laplacian} degenerates into the traditional Laplacian or Gaussian distribution when $\gamma$ equals $1$ or $2$.
Dilip Krishnan and Rob Fergus~\cite{krishnan2009fast} pointed out that the hyper-Laplacian with $\gamma = \sfrac{2}{3}$ is a better
model of image gradients than a Laplacian or a Gaussian. Here, we list the quantitative performance of our BSRDM 
under different settings of $\gamma$ in Table~\ref{tab:hyper-gamma}. It can be easily observed that BSRDM achieves the best
results when $\gamma$ equals to $\sfrac{2}{3}$, which is in accordance with the conclusion of Dilip Krishnan and Rob Fergus~\cite{krishnan2009fast}.
\begin{figure*}[t]
    \centering
    \includegraphics[scale=1.09]{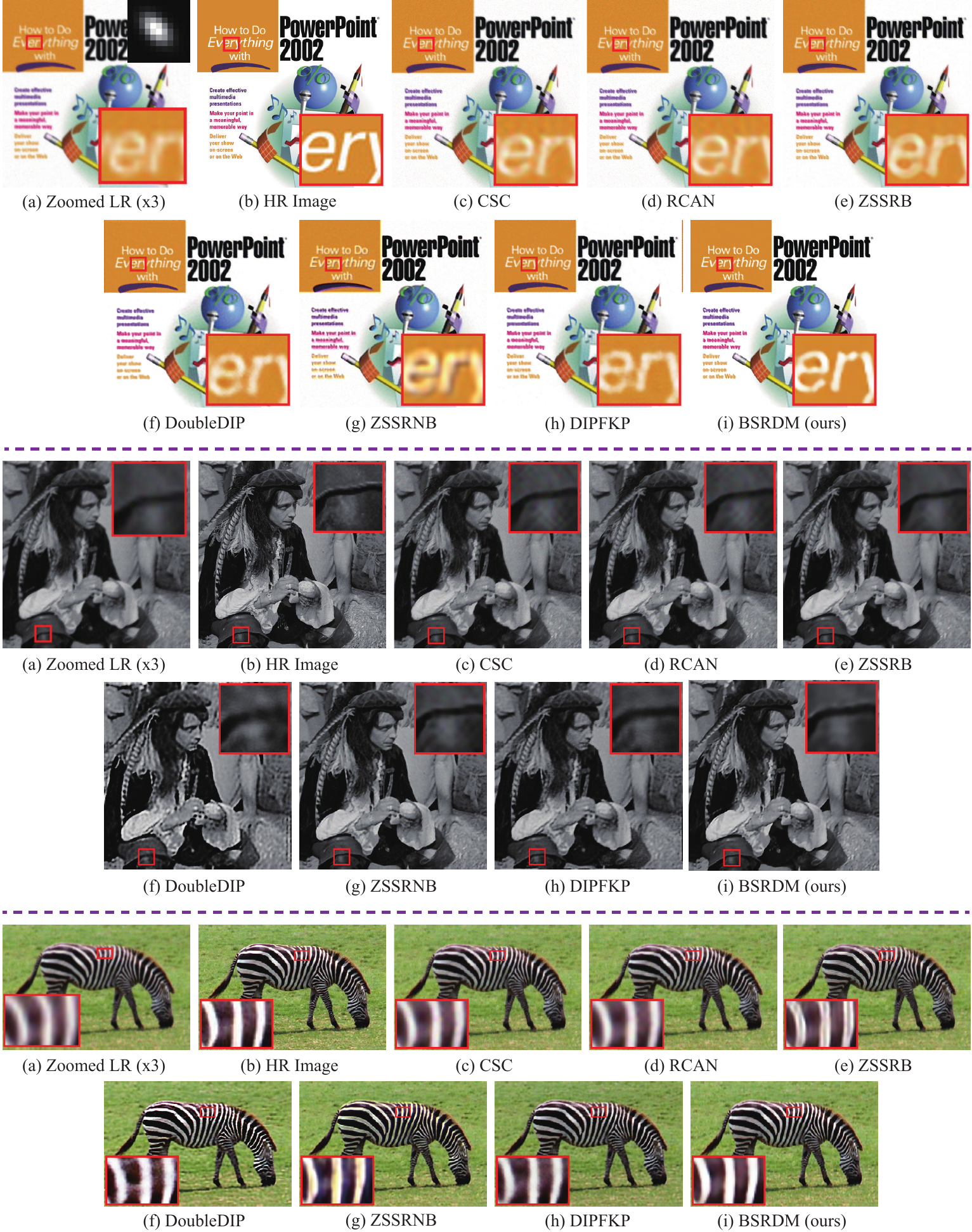}
    \vspace{-2mm}
    \caption{Visual super-resolution results of different methods for the degradation with Gaussian noise under scale factor 3. The blur kernel is 
            shown on the upper-right conner of the zoomed LR image.}
    \label{fig:syn_x3_gaussian_supp}
\end{figure*}
\begin{figure*}[t]
    \centering
    \includegraphics[scale=0.97]{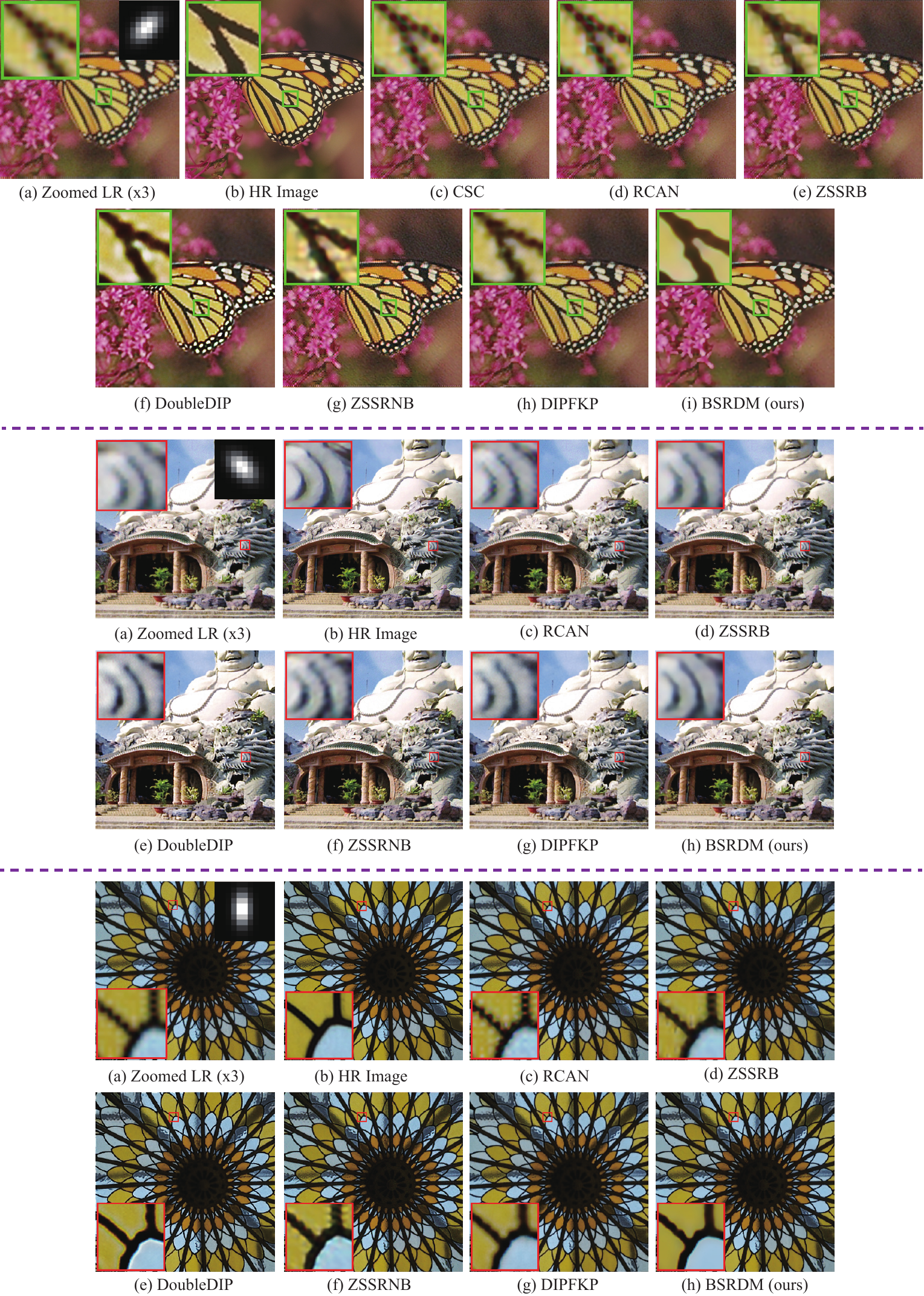}
    \vspace{-2mm}
    \caption{Visual super-resolution results of different methods for the degradation with camera sensor noise under scale factor 3. The blur kernel is 
             shown on the upper-right conner of the zoomed LR image. Note that due to the computer memory limitation, we cannot provide the super-resolution result of
             the method CSC for the second and third example in DIV2K100.}
    \label{fig:syn_x3_camera_supp}
\end{figure*}

\begin{table*}[t]
    \centering
    \caption{\footnotesize PSNR/SSIM/LPIPS results of different comparison methods on DIV2K100. All the results are averaged on six differen degradations with blur kernels
    as shown in Fig.~\ref{fig:kernels_six}. The best results are highlighted in \textbf{bold}. The \textcolor[gray]{0.5}{gray} results indicate unfair comparisons due to the mismatched degradations.}
    \label{tab:syn_DIV2K}
    \vspace{-3mm}
    \scriptsize
    \begin{tabular}{@{}C{1.8cm}@{}|@{}C{2.2cm}@{}|@{}C{1.5cm}@{}|@{}C{1.95cm}@{}|@{}C{1.95cm}@{}|@{}C{1.95cm}@{}|
                                                                 @{}C{1.95cm}@{}|@{}C{1.95cm}@{}|@{}C{2.05cm}@{}}
        \Xhline{0.8pt}
        \multirow{2}*{\makecell{Noise\\types}} & \multirow{2}*{Scale} & \multirow{2}*{Metrics} & \multicolumn{6}{c}{Methods} \\
        \Xcline{4-9}{0.4pt}
        &   &   &RCAN~\cite{zhang2018image} &ZSSR-B~\cite{shocher2018zero} &ZSSR-NB~\cite{shocher2018zero} &DoubleDIP~\cite{ren2020neural} &DIPFKP~\cite{liang2021flow} & BSRDM (ours)  \\
        \hline \hline
        \multirow{9}*{Case 1}  & \multirow{3}*{$\times$2}   & PSNR$\uparrow$      & \textcolor[gray]{0.5}{25.92}   & \textcolor[gray]{0.5}{26.00}   & \textbf{30.52}  & 25.17    & 27.38    & 29.07  \\
                               &                            & SSIM$\uparrow$      & \textcolor[gray]{0.5}{0.720}   & \textcolor[gray]{0.5}{0.734}   & \textbf{0.855}  & 0.689    & 0.749    & 0.800  \\
                               &                            & LPIPS$\downarrow$   & \textcolor[gray]{0.5}{0.343}   & \textcolor[gray]{0.5}{0.322}   & \textbf{0.284}  & 0.448    & 0.398    & 0.337  \\
        \Xcline{2-9}{0.4pt}
                               & \multirow{3}*{$\times$3}   & PSNR$\uparrow$      & \textcolor[gray]{0.5}{22.99}   & \textcolor[gray]{0.5}{23.13}   & 27.18           & 22.05    & 26.68    & \textbf{28.22}  \\
                               &                            & SSIM$\uparrow$      & \textcolor[gray]{0.5}{0.598}   & \textcolor[gray]{0.5}{0.616}   & 0.766           & 0.579    & 0.718    & \textbf{0.769}  \\
                               &                            & LPIPS$\downarrow$   & \textcolor[gray]{0.5}{0.407}   & \textcolor[gray]{0.5}{0.397}   & 0.376           & 0.517    & 0.452    & \textbf{0.373}  \\
        \Xcline{2-9}{0.4pt}
                               & \multirow{3}*{$\times$4}   & PSNR$\uparrow$      & \textcolor[gray]{0.5}{21.16}   & \textcolor[gray]{0.5}{21.43}   & 26.85           & 20.17    & 25.89    & \textbf{27.20}  \\
                               &                            & SSIM$\uparrow$      & \textcolor[gray]{0.5}{0.526}   & \textcolor[gray]{0.5}{0.548}   & \textbf{0.736}  & 0.514    & 0.696    & 0.732           \\
                               &                            & LPIPS$\downarrow$   & \textcolor[gray]{0.5}{0.467}   & \textcolor[gray]{0.5}{0.462}   & 0.423           & 0.546    & 0.474    & \textbf{0.414}  \\
        \hline \hline
r       \multirow{9}*{Case 2}  & \multirow{3}*{$\times$2}   & PSNR$\uparrow$      & \textcolor[gray]{0.5}{25.49}   & \textcolor[gray]{0.5}{25.69}   & 27.72           & 24.88    & 27.21    & \textbf{28.14}  \\
                               &                            & SSIM$\uparrow$      & \textcolor[gray]{0.5}{0.689}   & \textcolor[gray]{0.5}{0.708}   & 0.761           & 0.685    & 0.748    & \textbf{0.779}  \\
                               &                            & LPIPS$\downarrow$   & \textcolor[gray]{0.5}{0.415}   & \textcolor[gray]{0.5}{0.397}   & 0.397           & 0.460    & 0.415    & \textbf{0.385}  \\
        \Xcline{2-9}{0.4pt}
                               & \multirow{3}*{$\times$3}   & PSNR$\uparrow$      & \textcolor[gray]{0.5}{22.77}   & \textcolor[gray]{0.5}{22.91}   & 25.71           & 21.69    & 26.16    & \textbf{26.84}  \\
                               &                            & SSIM$\uparrow$      & \textcolor[gray]{0.5}{0.580}   & \textcolor[gray]{0.5}{0.599}   & 0.702           & 0.566    & 0.698    & \textbf{0.730}  \\
                               &                            & LPIPS$\downarrow$   & \textcolor[gray]{0.5}{0.497}   & \textcolor[gray]{0.5}{0.480}   & 0.470           & 0.541    & 0.492    & \textbf{0.401}  \\
        \Xcline{2-9}{0.4pt}
                               & \multirow{3}*{$\times$4}   & PSNR$\uparrow$      & \textcolor[gray]{0.5}{21.16}   & \textcolor[gray]{0.5}{21.24}   & 25.10           & 20.06    & 25.10    & \textbf{25.71}  \\
                               &                            & SSIM$\uparrow$      & \textcolor[gray]{0.5}{0.519}   & \textcolor[gray]{0.5}{0.538}   & 0.672           & 0.503    & 0.660    & \textbf{0.685}  \\
                               &                            & LPIPS$\downarrow$   & \textcolor[gray]{0.5}{0.551}   & \textcolor[gray]{0.5}{0.540}   & 0.517           & 0.582    & 0.535    & \textbf{0.509}  \\
        \Xhline{0.8pt}
    \end{tabular}
\end{table*}

\subsection{Limitations}
Figure~\ref{fig:real_limit} displays a real super-resolution example, in which the LR image is heavily corrupted by camera sensor noise. It can be easily observed that the current SotA
method DIPFKP cannot handle such case with complicated real noise, its recovered result contains obvious artifacts. On the contrary, the proposed BSRDM is able to remove most of the noise
and obtains clean super-resolved HR image. Even though achieving superior performance, BSRDM still has two major limitations. Firstly, the recovered image of BSRDM is smooth, since the $L_2$
loss function (see Eq.~\eqref{eq:Q_fun_minimization}) and the hyper-Laplacian prior on image gradients in it both favor smoothing the generated HR image. Secondly, BSRDM cannot hallucinate more image textures
that not exists in the observed LR image, e.g., hairs of the dog in Fig.~\ref{fig:real_limit}, and is thus inferior to the GAN-based methods~\cite{wang2021real,wang2021real} on this point. In
the future, it might be expected to develop more powerful image priors specifically to overcome these limitations.

\subsection{Experiments on the Real Data}
\subsubsection{More Visual Results}\label{subsubsec:visual_real_exp}
Figure \ref{fig:real_x4_supp} displays three more visual super-resolution results on RealSRSet~\cite{zhang2021designing} with scale factor 4. In the first (top row) and second (middle row) examples,
the LR image is with obvious camera sensor noise.
The comparison methods cannot deal with such degradation with complicated real noises, while our BSRDM is able to remove most of the noises, indicating the effectiveness of the proposed
non-i.i.d. noise modeling method. In the third example (bottom row), it can be easily seen that the recovered HR image by BSRDM is with sharper clearer details.

\subsubsection{Disscussion on the Non-reference Metrics}
Since the ground-truth for the RealSRSet~\cite{zhang2021designing} is not available, three non-reference metrics (i.e., NIQE~\cite{mittal2012making}, NRQM~\cite{Ma2017} and PI~\cite{blau20182018}) are considered as quantitative
evaluation. As shown in Table~\ref{tab:non-refer}, BSRDM and the current SotA method DIPFKP~\cite{liang2021flow} both fail to achieve promising results. However, in Fig.~\ref{fig:real_x4_supp} and
Fig.~\ref{fig:real_x4_maintext}, we can easily observed that the recovered results by BSRDM is evidently better than other comparison methods.
We argue that these non-reference metrics are not consistent with our perceptual visual system. In the future work, we will make our best effort to
develop more rational non-reference metric to match with and facilitate current researches on SISR.

\subsection{Experiments on the Synthetic Data}
In Table~\ref{tab:syn_DIV2K}, we list the performance comparisons of different methods on the dataset DIV2K100~\cite{agustsson2017ntire}. Note that, due to the computer memory limitation,
we cannot give the results of CSC~\cite{gu2015convolutional} in Table~\ref{tab:syn_DIV2K}. It can be
easily observed that the proposed BSRDM illustrates obvious superiorities than the comparison methods, which is consistent with that on Set14 in Table~\ref{tab:syn_set14},
Furthermore, we display more visual results of different methods on the synthetic data
sets in Fig.~\ref{fig:syn_x3_gaussian_supp} (Gaussian noise) and Fig.~\ref{fig:syn_x3_camera_supp} (camera sensor noise).

\end{document}